\documentclass[10pt,twocolumn,letterpaper]{article}

\usepackage{cvpr}
\usepackage{times}
\usepackage{epsfig}
\usepackage{graphicx}
\usepackage{amsmath}
\usepackage{amssymb}
\usepackage{array}
\usepackage{multirow,bigdelim}
\usepackage{rotating}
\usepackage{hhline}
\usepackage{xcolor}
\usepackage{afterpage}
\usepackage{geometry}
\newcommand\crule[3][black]{\textcolor{#1}{\rule{#2}{#3}}}
\def\labelvars{\mathbf{f}}
\def\labelvar{f}
\def\binvars{\mathbf{x}}
\def\binvar{x}

\def\pixelset{\mathcal{P}}
\def\neighbset{\mathcal{N}}
\def\shapeedges{\mathcal{E}}
\def\labelset{\mathcal{L}}
\def\labelsetindex{k}
\def\unary{D}
\def\pottspairwise{V}
\def\shapepairwise{S}

\def\Hhogs{H}
\def\coneangle{\theta}
\def\a{$\alpha$}
\cvprfinalcopy 

\def\cvprPaperID{1563} 

\ifcvprfinal\fi
\begin{document}

\title{$A$-expansion for multiple ``hedgehog'' shapes}

\author{Hossam Isack \hspace{6ex}  Yuri Boykov  \hspace{6ex} Olga Veksler\\
Computer Science \\
University of Western Ontario, Canada \\
{\tt\small   habdelka@csd.uwo.ca \hspace{1ex} yuri@csd.uwo.ca \hspace{1ex}olga@csd.uwo.ca}
}

\pagestyle{myheadings}
\setlength{\headsep}{0.5in}
\markright{\small  H. Isack, Y. Boykov, O. Veksler,  arXiv, Feb. 2, 2016}
\setcounter{page}{1} 

\maketitle
\thispagestyle{myheadings}

\vspace{0.5ex}
\begin{abstract}
\vspace{0.5ex}
Overlapping colors and cluttered or weak edges are common segmentation problems requiring additional
regularization. For example, {\em star-convexity} is popular for interactive single object
segmentation due to simplicity and amenability to exact graph cut optimization.
This paper proposes an approach to multi-object segmentation where objects could
be restricted to separate ``hedgehog'' shapes. We show that $\alpha$-expansion moves
are submodular for our multi-shape constraints. Each ``hedgehog'' shape has its surface normals constrained
by some vector field, e.g.~gradients of a distance transform for user scribbles.
Tight constraint give an extreme case of a shape prior enforcing skeleton
consistency with the scribbles. Wider cones of allowed normals gives more relaxed hedgehog shapes.
A single click and $\pm 90^{\circ}$ normal orientation constraints reduce our hedgehog prior to star-convexity.
If all hedgehogs come from single clicks then our approach defines multi-star prior.
Our general method has significantly more applications than standard one-star segmentation.
For example, in medical data we can separate multiple non-star organs with similar
appearances and weak or noisy edges.
\end{abstract}

\vspace{0.5ex}
\section{Introduction}
\begin{figure}[t]
\begin{center}
\begin{tabular}{c@{\hspace{1.5ex}}c}
\includegraphics[height=0.41\linewidth]{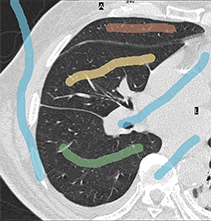} &
\includegraphics[height=0.41\linewidth]{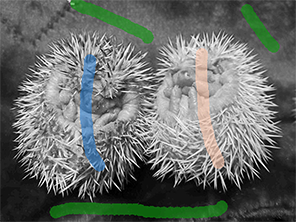}\\[-1ex]
\multicolumn{2}{c}{\small two examples of images with seeds (medical and photo)} \\
\includegraphics[height=0.41\linewidth]{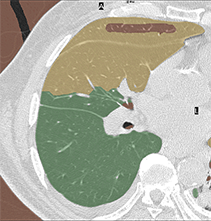} &
\includegraphics[height=0.41\linewidth]{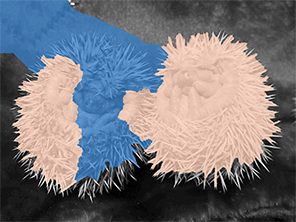}\\[-1ex]
\multicolumn{2}{c}{\small multi-object segmentation using Potts model} \\
\includegraphics[height=0.41\linewidth]{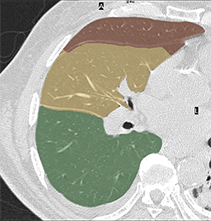} &
\includegraphics[height=0.41\linewidth]{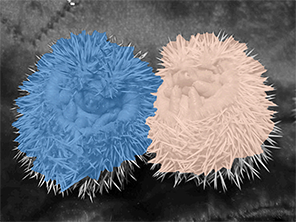}\\[-1ex]
\multicolumn{2}{c}{\small multi-object segmentation adding our {\em hedgehog} shapes prior}
\end{tabular}
\end{center}
   \caption{{\em Hedgehog shapes} prior for multi-object segmentation.}
\label{fig:teaser}
\end{figure}

\vspace{0.5ex}
\label{sec:intro}
Distinct intensity appearances and smooth contrast-aligned boundaries are standard segmentation cues.
However, in most real applications of image segmentation there are multiple objects of interest with
similar or overlapping color appearances. Intensity edges also could be cluttered or weak. These common
practical problems require additional regularization, as illustrated in the second row of Figure \ref{fig:teaser}.

There are multiple methodologies for enforcing shape regularity or shape priors.
For example, Statistical Shape Models (SSM) \cite{vu2008shape,cremers2008shape,ConvexMultiRegionICCV2011}
and Deformable Shape Models (DSM) \cite{pizer2003deformable}
differ in their shape space representation and their distance measures between a given segmentation and
the learned shape space. SSM applies principal component analysis to a training dataset for fitting
the shape space distribution represented by a mean shape and the modes of greatest variance. Any given segmentation
could be penalized based on how well it aligns to this shape space \cite{vu2008shape}
or it could be restricted to the learned shape space \cite{cremers2008shape}.
M-rep \cite{pizer2003deformable} is a coarse-to-fine discrete DSM approach. In contrast to basic
user-scribbles like in our method, M-reps requires detailed user inputs defining a figural shape model
for each segment. They also need training data to estimate their model parameters.
SSM and DSM assume a fixed shape topology, which is often violated by specific problem instances,
\eg~lesions, tumors, or horse-shoe kidneys \cite{siddiqi2008medial}.

Our paper proposes a simple and sufficiently
general shape regularization constraint that could be easily integrated into standard MRF methods for segmentation.
Shape priors have been successfully used in binary graph cut segmentation \cite{KB:ICCV05,olga:ECCV08,GeodesicStarCVPR10}.
While our ``hedgehog'' shape prior is a generalization of the popular {\em star-convexity} constraint \cite{olga:ECCV08}
with several merits over previous extensions \cite{KB:ICCV05,GeodesicStarCVPR10}, our main contribution is
a {\em multi-hedgehog} prior in the context of multi-object segmentation problems.

We observe that similarity between object appearances and edge clutter are particularly problematic in
larger multi-label segmentation problems, e.g.~in medical imaging. Our multi-hedgehog prior is fairly flexible,
has efficient optimizers, and shows significant potential in resolving very common
ambiguities in multi-label segmentation problems, see Fig.\ref{fig:teaser} (last row).
Our general multi-object segmentation framework allows to enforce
``hedgehog'' shape prior for any of the objects. The class of all possible hedgehog priors is sufficiently representative
yet each specific hedgehog constraint offers sufficient regularization to address color overlap and weak/cluttered edges.
One extreme case of our prior is closely related to the standard {\em star} shape prior \cite{olga:ECCV08}. The other extreme case
allows shapes with restricted {\em skeletons} \cite{Mreps:IJCV03,SiddiqiPizer:2008}.

The main contribution of our work is a practical and efficient way to combine
distinct shape priors for segments in popular {\bf multi-label} MRF framework \cite{BVZ:PAMI01}.
Our work also allows to extend previous multi-surface graph cut methods \cite{sonka:PAMI06,DB:ICCV09}.
For example, \cite{sonka:PAMI06} compute multiple nested segments using one fixed {\em polar} grid
defined by some non-overlapping {\em rays}. Besides particular image discretization,
these rays introduce two constraints: one star-like shape constraint shared by the nested segments
and a smoothness constraint penalizing segment boundary jumping between adjacent
rays. In contrast, our method defines independent shape constraints
for each segment. Similarly to \cite{KB:ICCV05}, shape normals are constrained by arbitrary vector fields,
rather than non-overlapping rays \cite{sonka:PAMI06} or trees \cite{olga:ECCV08,GeodesicStarCVPR10}.
Our use of Cartesian grid allows to enforce standard boundary length smoothness \cite{BK:ICCV03}.
While this paper is focused on a Potts model with distinct shape constrains, hedgehog shapes can be easily combined
with inter-segment {\em inclusion} or {\em exclusion} constraints \cite{DB:ICCV09}.
The use of distinct (not necessarily nested) shape priors extends the range of applications in \cite{sonka:PAMI06}.

Our Potts framework optimization algorithm is closely comparable with a special non-submodular case of \cite{DB:ICCV09} with
exclusion constraints.  While we use independent shape constrains for each segment, these are easy to integrate into
each layer in \cite{DB:ICCV09}. More importantly, instead of binary multi-layer formulation with non-submodular potentials,
we use multi-label formulation amenable to $\alpha$-expansion. Besides memory savings,
our approach solves the non-submodular segmentation problem with a guaranteed approximation quality bound.
Section \ref{sec:related_graphcuts} discusses relations to \cite{sonka:PAMI06,DB:ICCV09} in more details.


{\bf Overview of contributions:} We  propose a new multi-label segmentation model
and the corresponding optimization algorithm.  Our contributions are summarized below.
\begin{itemize}
\itemsep0em
\item {\em hedgehog shape constraint} - a new flexible method for segment regularization based on simple and intuitive user interactions.
\item new multi-object segmentation energy with {\em multi-hedgehog} shape priors.
\item we provide an extension of $\alpha$-expansion moves \cite{BVZ:PAMI01} for the proposed energy.
\item experimental evaluation showing how our multi-object segmentation method solves problematic
cases for the standard Potts model \cite{BVZ:PAMI01}.
\end{itemize}
The rest of the paper is organized as follows. Section \ref{sec:hhogconst} defines our hedgehog shape prior for a simpler case of
binary segmentation of one object. We discuss its properties and show how it can be globally optimized with $s/t$ graph cuts.
Section \ref{sec:segenergy} defines multi-hedgehog shape constraint in the context of multi-label MRF segmentation and proposes
an extension of $\alpha$-expansion optimization algorithm. Our experiments in Section \label{sec:experiments} includes
multi-object segmentation of real photos and 3D multi-modal medical data.

\section{Hedgehog shape constraint for one object}    \label{sec:hhogconst}

This section describes our {\em hedgehog} shape prior for a single object
in case of binary segmentation. Section  \ref{sec:segenergy} describes
a more general {\em multi-hedgehog} segmentation prior where multiple objects
can have separate hedgehog constraints. While {\em multi-hedgehog} prior
helps in a much wider range of problems, e.g.~in medical imaging,
binary segmentation with one ``hedgehog'' is easier to start from and it
has merits on its own. In particular, single-hedgehog prior generalizes popular {\em star-convexity}
\cite{olga:ECCV08} differently from other related methods \cite{KB:ICCV05,GeodesicStarCVPR10}
in binary segmentation.
\begin{figure}[t]
\begin{center}
\begin{tabular}{@{\hspace{-1ex}}c@{\hspace{1ex}}c}
\includegraphics[width=0.5\linewidth]{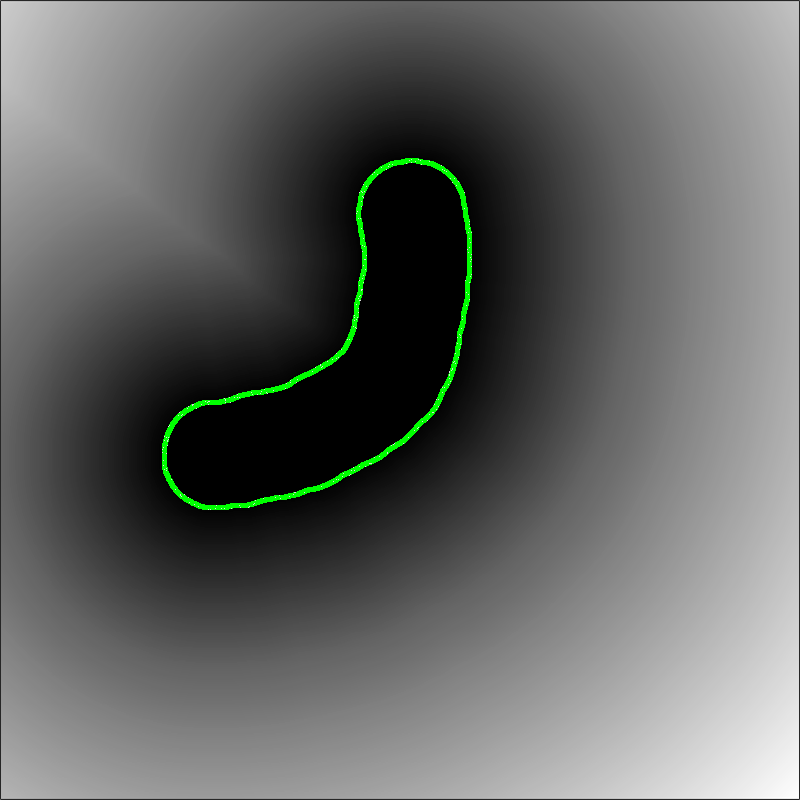} &
\includegraphics[width=0.5\linewidth]{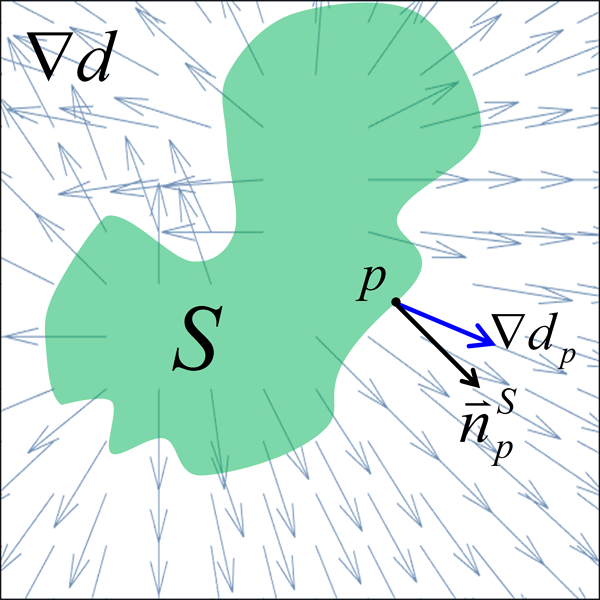} \\
(a) scribble's distance map $d$ & (b) constraint $\angle({\bar n}^S_p,\nabla d_p)\leq\theta$
\end{tabular}
\end{center}
   \caption{{\em Hedgehog prior} for segment $S$. (a) User-scribble or shape template define a (signed) distance map $d$.
(b): Orientations of surface normals ${\bar n}^S_p$ for $S$ are constrained by
$\angle({\bar n}^S_p,\nabla d_p)\leq\theta$. }
\label{fig:continuous}
\end{figure}

Similarly to {\em star} prior \cite{olga:ECCV08}, hedgehog prior could be defined interactively.
Instead of a single click in the star center, hedgehog shape allows an arbitrary scribble
roughly corresponding to its {\em skeleton}. Hedgehog can also be defined by
an approximate user-defined outline of a desired shape or by a shape template. In any case,
such scribble, outline, or template define the corresponding (signed) {\em distance transform} or
{\em distance map} $d:\Omega\rightarrow{\mathcal R}$ and a field of its gradients $\nabla d$, as illustrated in
Fig.\ref{fig:continuous}. Our {\em hedgehog constraint} for segment $S$ is defined
by vector field $\{\nabla d_p\;|\;p\in\Omega\}$ and angular threshold $\theta$ restricting
orientations of surface normals ${\bar n}^S_p$ at any point $p$ on the boundary of $S$ to a cone
\begin{equation}\label{eq:h_theta}
C_\theta(p):\;\;\;\;\; \angle({\bar n}^S_p,\nabla d_p)\leq\theta \;\;\;\;\;\;\;\;\forall p \in\partial S
\end{equation}
assuming gradient $\nabla d_p$ is defined at $p$.
More generally,  {\em hedgehog constraint} for segment $S$ could be defined by any given vector field
$\{{\bar v}_p\;|\;p\in\Omega\}$ defining preferred directions for surface normals. Similarly to \cite{KB:ICCV05},
we can use {\em dot product} to define allowed normals cones
$C_\theta(p):\;\langle{\bar n}^S_p,{\bar v}_p\rangle\geq \tau$ where width varies depending
on the magnitude of ${\bar v}_p$. In case ${\bar v}_p=\nabla d_p$
this constraint reduces to \eqref{eq:h_theta} for $\tau = cos(\theta)$ since $|\nabla d_p|=1$
at all points where gradient $\nabla d_p$ exists.

\subsection{Single hedgehog properties}
\label{sec:hhogprop}
\begin{figure}[t]
\begin{center}
\begin{tabular}{@{\hspace{-2ex}}c@{\hspace{-1ex}}c}
\includegraphics[width=0.55\linewidth]{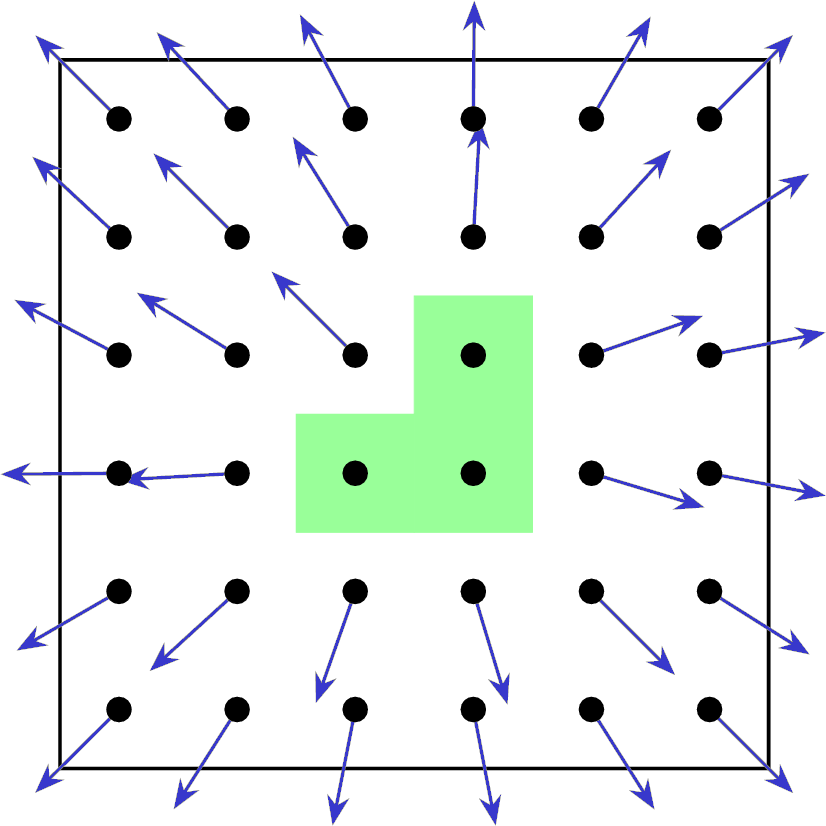} &
\includegraphics[width=0.55\linewidth]{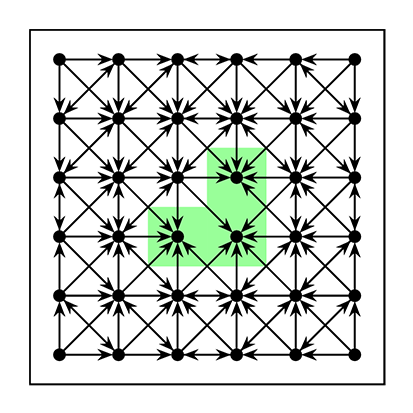} \\
(a) gradients $\nabla d$ & (b) graph edges $\shapeedges(0)$\\
\includegraphics[width=0.55\linewidth]{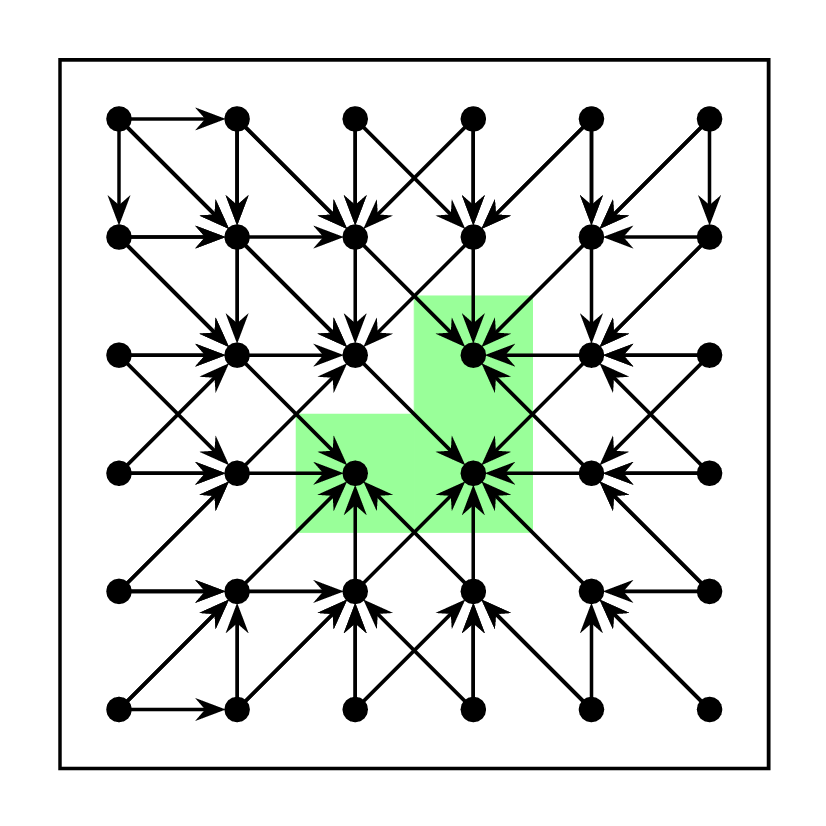} &
\includegraphics[width=0.55\linewidth]{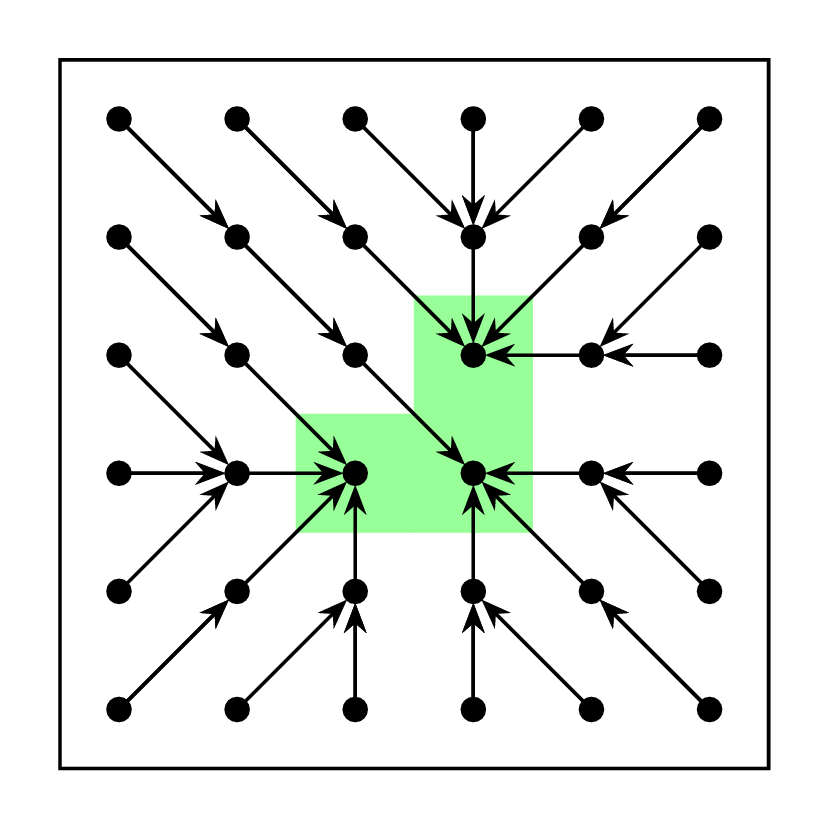} \\
(c) graph edges $\shapeedges(\frac{\pi}{4})$ & (d) graph edges $\shapeedges(\frac{3\pi}{8})$
\end{tabular}
\end{center}
\caption{Hedgehog constraint \eqref{eq:h_theta} for user seeds (green) and the corresponding
distance map gradients $\nabla d$ in (a) is approximated by infinity cost directed edges $\shapeedges(\theta)$ in (b-d)
selected as in Fig.\ref{fig:twocones}.}
\label{fig:discreteexample}
\end{figure}

Even a single hedgehog shape prior discussed in this section could be useful in practice.
For example, if $\theta = \pi/2$ it closely approximates a popular star convexity \cite{olga:ECCV08}
in case of a single click. However, our formulation uses locally defined constraints, which can be
approximated by a simple rule for selecting local edges, see Section \ref{sec:hhogviagc}. Unlike \cite{olga:ECCV08},
we do not enforce a global tree structure, see Fig.\ref{fig:discreteexample}(b).  Also, like \cite{KB:ICCV05,GeodesicStarCVPR10}
hedgehog prior allows a much larger variability of shapes for scribbles different from a point.
In our case, a scribble defines a rough {\em skeleton} of a shape. For example, for smaller values of $\theta$
our cone constraints \eqref{eq:h_theta} give a tighter alignment of surface normals with vectors
$\nabla d_p$ forcing the segment boundary to closely follow the level-sets of the scribble's distance map $d$,
see Fig.\ref{fig:varyingtheta}. In the limit, this implies consistency of segment's skeleton with the skeleton of the given
scribble, outline, or template.

\subsection{Single hedgehog via graph cuts} \label{sec:hhogviagc}

\begin{figure}[t]
\begin{center}
\begin{tabular}{@{\hspace{-2ex}}c@{\hspace{0.5ex}}c}
\includegraphics[width=0.51\linewidth]{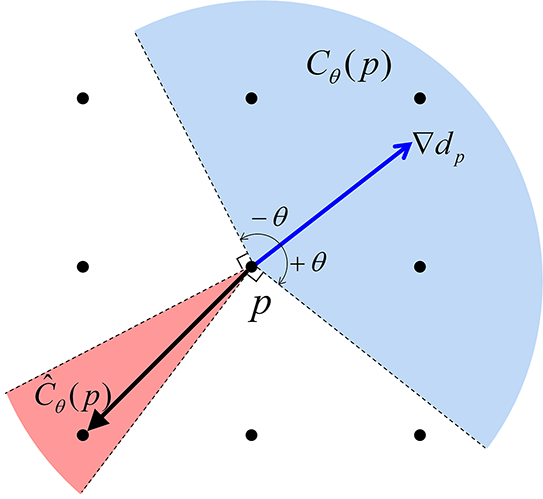} &
\includegraphics[width=0.51\linewidth]{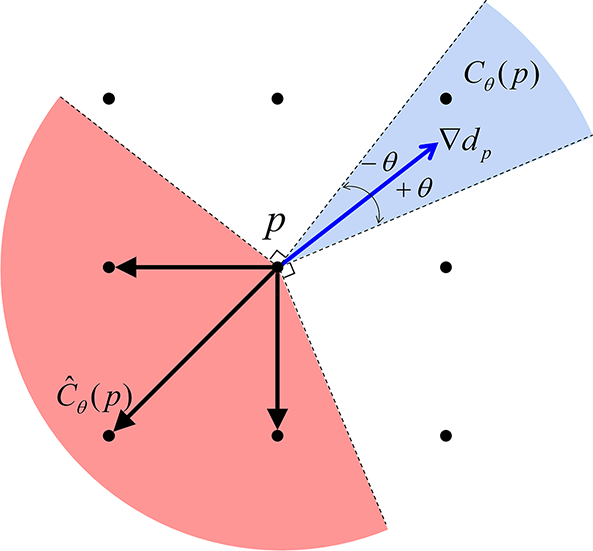}\\
(a) wide cone of normals $C_{\theta}$ &  (b) tight cone of normals $C_{\theta}$
\end{tabular}
\end{center}
\caption{Approximating {\em hedgehog constraint} \eqref{eq:h_theta} at grid node $p$.
Cone $C_{\theta}$ of allowed surface normals (blue) is enforced by $\infty$ cost directed edges $(pq)$
in the corresponding {\em polar cone} ${\hat C}_{\theta}$ (red).}
\label{fig:twocones}
\end{figure}

\begin{figure*}
\begin{center}
\begin{tabular}{ccc}
\includegraphics[width=0.3\linewidth]{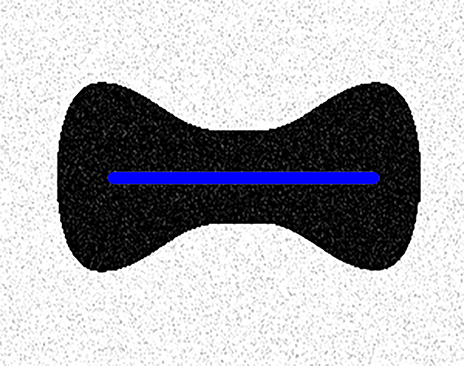} &
\includegraphics[width=0.3\linewidth]{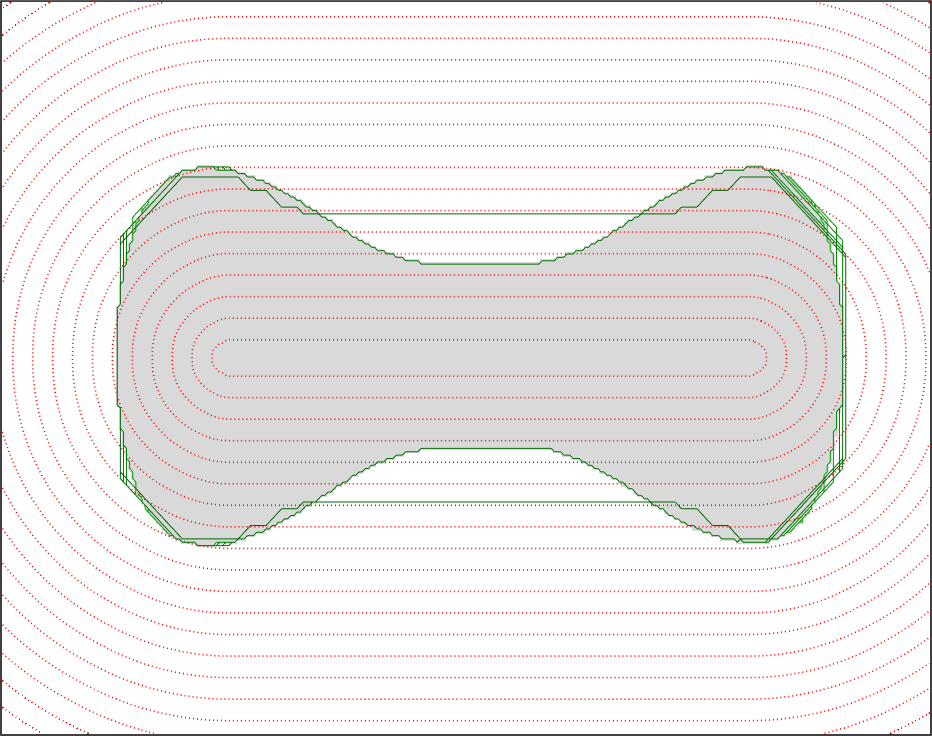} &
\includegraphics[width=0.3\linewidth]{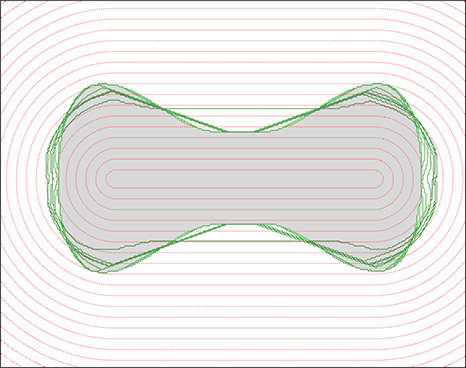} \\ \\
\includegraphics[width=0.3\linewidth]{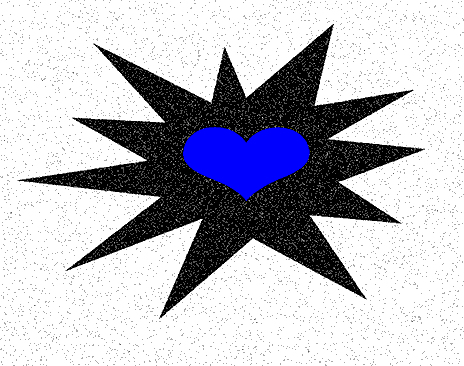} &
\includegraphics[width=0.3\linewidth]{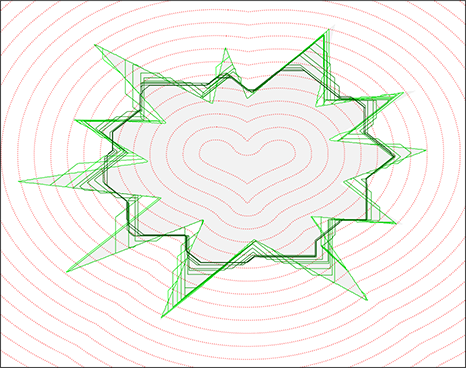} &
\includegraphics[width=0.3\linewidth]{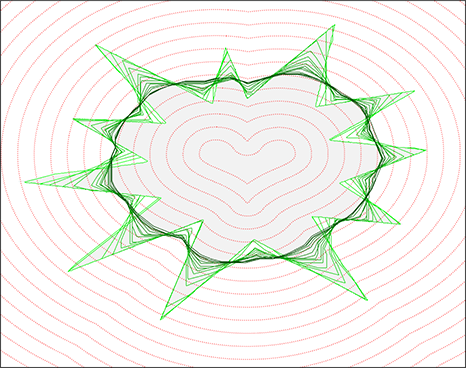} \\ \\
\includegraphics[width=0.3\linewidth]{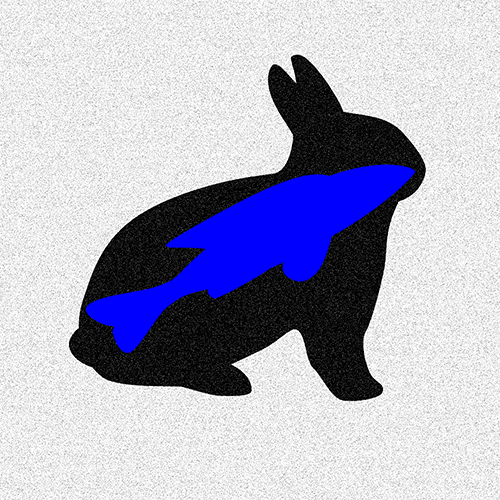} &
\includegraphics[width=0.3\linewidth]{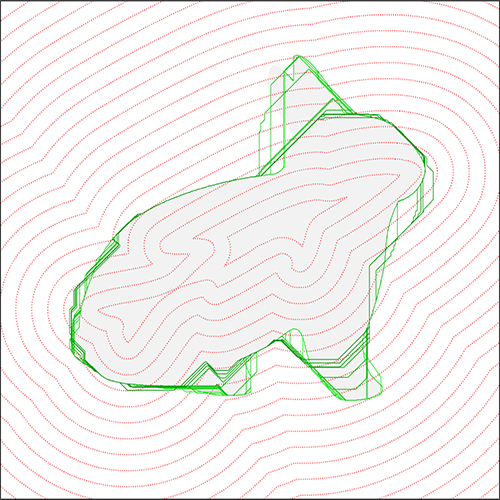} &
\includegraphics[width=0.3\linewidth]{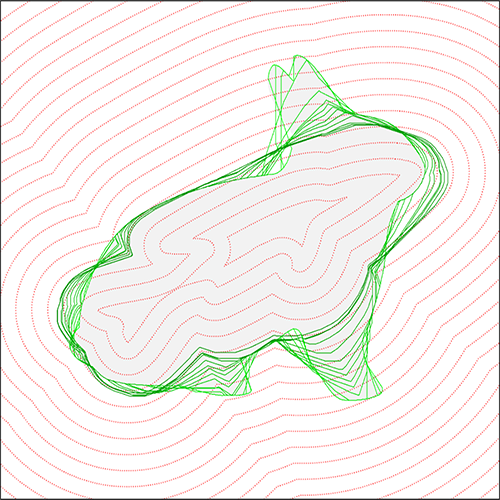}\\ \\
(a) image and user scribble (blue) & (b) 8-neighborhood & (c) 32-neighborhood
\end{tabular}
\end{center}
\caption{Single {\em hedgehog shape constraint} \eqref{eq:h_theta} for different parameters $\theta$.
User scribbles in (a) define distance map $d$ and, consequently, vector field $\nabla d$ in \eqref{eq:h_theta}.
Red contours in (b,c) are level sets of distance map $d$. Green contours show optimal graph cut segmentation
(Sec.\ref{sec:hhogviagc}) using hedgehog constrains with different values of parameter $\theta$ in the range $[0,\frac{\pi}{2}]$.
In these synthetic examples the object boundary nearly satisfies our hedgehog constraint for $\theta =\frac{\pi}{2}$.
However, smaller parameters $\theta$ (darker green contours) correspond to tighter constraints for segment normals,
see Fig.\ref{fig:twocones}, forcing object and background segments to deviate from their given color models.
As $\theta$ approaches $0$, constraint \eqref{eq:h_theta} closely aligns segments with level sets of $d$ (skeleton consistency).
Discretization artifacts decrease for larger $\neighbset$, see (c).}
\label{fig:varyingtheta}
\end{figure*}

We show an approximation for {\em hedgehog constraint} \eqref{eq:h_theta} for object $S$
in the context of binary N-dimensional image segmentation via graph cuts \cite{BJ:01}.
All cone constraints \eqref{eq:h_theta} for any given $\theta$ and distance map gradients $\nabla d$,
see Fig.\ref{fig:discreteexample}(a), correspond to a certain set of infinity cost directed edges,
see Fig.\ref{fig:discreteexample}(b-d).
For example, consider cone of allowed surface normals $C_{\theta}(p)$ at some point $p$ illustrated in
Fig.\ref{fig:twocones} for two different values of parameter $\theta$. It is easy to see that
a surface/boundary of segment $S$ passing at $p$ has normal ${\bar n}^S_p \in C_\theta(p)$ iff
this surface does not  cross the corresponding {\em polar cone}
\begin{equation} \label{eq:polar_cone}
{\hat C}_{\theta}(p) \;:=\; \{(py)\,|\, \langle(py),(pz)\rangle \leq 0\;\;\forall z\in C_{\theta}(p)\}.
\end{equation}
This reformulation of our hedgehog constraint \eqref{eq:h_theta} is easy to approximate via graph cuts
by setting infinity cost to all directed edges adjacent to $p$ whose directions agrees with polar cone
${\hat C}_{\theta}(p)$, see Fig.\ref{fig:twocones}.
To avoid clutter, the figure only shows such directed edges $(pq)\in {\hat C}_{\theta}(p)$ starting at $p$,
but one should also include similarly oriented directed edges
$(qp)\in -{\hat C}_{\theta}(p):= \{(yp)\,|\, \langle(yp),(pz)\rangle \leq 0\;\;\forall z\in C_{\theta}(p)\} $
pointing to $p$. The set of all directed graph edges consistent with local polar cones orientations,
see Fig.\ref{fig:discreteexample}(b-d), is
\begin{equation} \label{eq:e_theta}
\!\!\!\!\shapeedges(\theta)\!=\!\{(pq)\in \neighbset \; | \; (pq) \in \hat{C}_{\theta}(p)\;or\; (pq) \in -\hat{C}_{\theta}(q) \}.
\end{equation}
Obviously, hedgehog constraints are better approximated by larger neighborhood systems $\neighbset$,
\eg 32-neighborhood works better than 8-neighborhood, see Fig.\ref{fig:varyingtheta}(b,c).

\begin{figure*}[t]
\begin{center}
\begin{tabular}{ccc}
\includegraphics[height=0.3\linewidth]{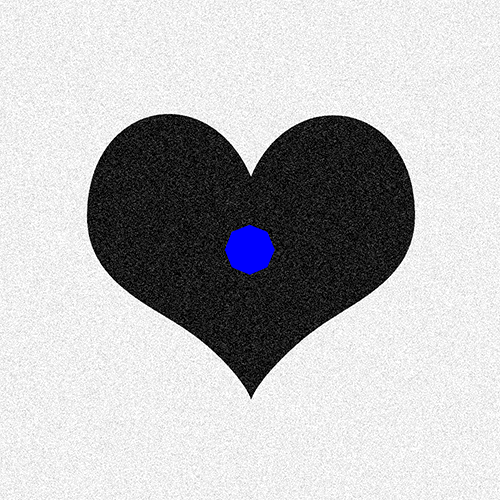} &
\includegraphics[height=0.3\linewidth]{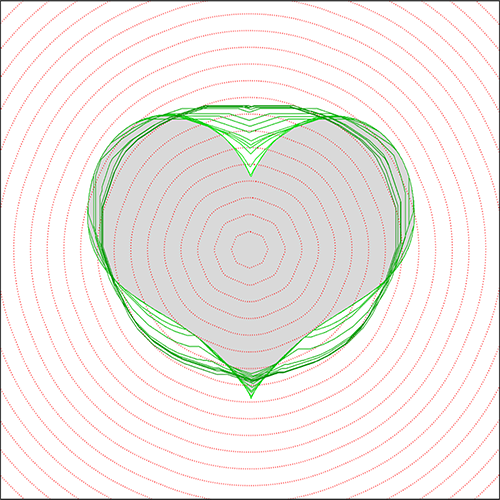} &
\includegraphics[height=0.3\linewidth]{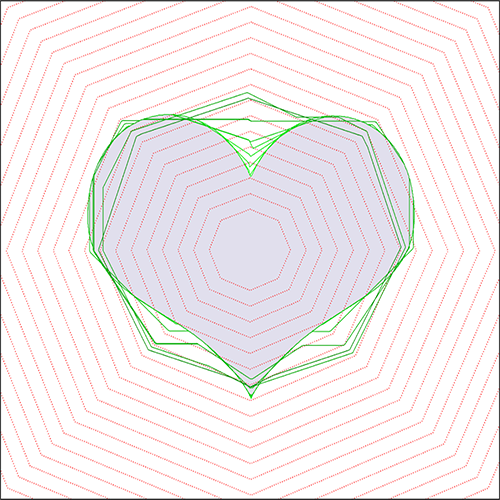} \\
{\small (a) image and user scribble (blue)} & {\small (b) Euclidean Distance Transform gradient} & {\small (c) synthetic vector field}
\end{tabular}
\end{center}
\caption{Single {\em hedgehog shape constraint} \eqref{eq:h_theta} for different parameters $\theta$ and different vector fields.
(a) image and scribble (octagon shown in blue). (b-c) show our results for varying $\theta$, ground truth (shown in gray) and the vector field used
to enforce hedgehog constraints (shown in blue). In (b) the vector field is the Euclidean Distance Transform gradient of the user-scribble and in (c) it is the gradient of a function
where the level-sets are scaled versions of the scribble. Notice how the used vector field affects the segmentation for the same value of $\theta$ in
(b) and (c).
}
\label{fig:varyingthetaAndVectorField}
\end{figure*}

The used vector field has a direct effect on the set of allowed shapes when varying $\theta$. Figure~\ref{fig:varyingthetaAndVectorField} shows the segmentation result for varying $\theta$ for two different vector fields on the same synthetic example.


It is easy to see that set \eqref{eq:e_theta} of infinity cost edges corresponds to a {\em submodular} pairwise energy
approximating hedgehog shape constraint \eqref{eq:h_theta} for binary labeling $\labelvars=\{\labelvar_p\}$
representing segment $S=\{p\;|\;\labelvar_p=1\}$
\begin{equation} \label{eq:h}
h_\theta(\labelvars) \;=\; \sum_{(pq)\in \shapeedges(\theta) } w_{\infty}\cdot[\labelvar_p=1,\labelvar_q = 0]
\end{equation}
where $w_{\infty}$ is an infinitely large  scalar.

\section{Multi-hedgehog segmentation energy}  \label{sec:segenergy}
Given a set of pixels $\pixelset$, neighborhood system $\neighbset$, and labels $\labelset$ our multi-labeling segmentation energy is
\begin{equation} \label{eqn:OurMainEnergy}
E(\labelvars) = \overbrace{\sum_{p \in \pixelset} \unary_p(\labelvar_p)}^\text{\small {\em data}} + \overbrace{\lambda\!\!\sum_{pq \in \neighbset}\pottspairwise_{pq}(\labelvar_p,\labelvar_q)}^\text{\small {\em smoothness}} + \overbrace{\vphantom{\sum} \Hhogs_\coneangle(\labelvars)}^\text{\small {\em hedgehogs}}
\end{equation}
where $\labelvars=\{\labelvar_p \in \labelset |\; \forall p \in \pixelset \}$ is a labeling.

The first two terms, namely {\em data}  and {\em smoothness} terms, are widely used in computer vision, e.g.~\cite{BVZ:ICCV99,BJ:01,GrabCuts:SIGGRAPH04}.
The data term commonly referred to as the {\em regional} term as it measures how well pixels fit into their corresponding labels. To be specific,
$\unary_p(\labelvar_p$) is the penalty for assigning label $\labelvar_p$ to pixel $p$.
Similar to \cite{GrabCuts:SIGGRAPH04}, a label's probabilistic model, a Gaussian Mixture in our case, is found by fitting a probabilistic model to the seeds given by the user.

The smoothness term is a standard pairwise regularizer that
discourages segmentation discontinuities between neighboring pixels.
A discontinuity occurs whenever two neighboring pixels $pq\in \neighbset$ are assigned to different labels. In its simplest form,
$\pottspairwise_{pq}(\labelvar_p,\labelvar_q)= w_{pq}[\labelvar_p\neq\labelvar_q]$ where $[\;]$ is {\em Iverson bracket} and
$w_{pq}$ is a non-increasing function of the intensities at $p$ and $q$.
Also, $\lambda$ is a parameter that weights the importance of the smoothness term.

Third term, our contribution, is the {\em Hedgehog term}
\begin{equation} \label{eqn:shapepenalty}
\Hhogs_\coneangle(\labelvars)=\sum_{\labelsetindex \in \labelset} \sum_{\;(pq) \in \shapeedges_{\labelsetindex}(\coneangle)}\!\!\! w_{\infty}\;[\labelvar_p=k,\labelvar_q\neq k]
\end{equation}
where $w_{\infty}=\infty$. Those familiar with graph cuts may prefer to think of it as an $\infty$-cost arc from $p$ to $q$, thus prohibiting
any cut that satisfy $\labelvar_p=k$ and $\labelvar_q\neq k$.

The Hedgehog term is the sum of the Hedgehog constraints over all the labels and it guarantees that any feasible labeling\footnote{We use feasible (and not bound) because there is at least one trivial solution with finite cost. In practice, it is practical to assume that one of the labels, e.g.~background label, does not require enforcing shape constraints otherwise the problem could become over-constrained. One trivial solution is to label all pixels as background except those labeled by user scribbles.}, i.e.~$E(\labelvars)<\infty$, will result in a segmentation with surface normals respecting the orientation constraints \eqref{eq:h_theta}. Notice that \eqref{eqn:shapepenalty} reduces to \cite{olga:ECCV08} when $\theta=90^{\circ}$, $|\labelset|=2$ and the shape constraints are defined for only one of the labels by a single pixel.

\subsection{Expansion Moves}
\label{sec:expmoves}
In this section we will describe how to extend the binary expansion moves of \a-exp \cite{BVZ:PAMI01} to respect the shape constraints, and show that these moves are submodular.
The main idea of \a-exp algorithm is to maintain a current feasible labeling $\labelvars^{'}\!\!$, i.e.~$E(\labelvars^{'})\!\!<\!\!\infty$, and iteratively move to a better labeling until no improvements could be made.
To be specific at each iteration, a label $\alpha \in \labelset$ is chosen and variables $\labelvar_p$ for all $p \in \pixelset$ are given a binary choice $\binvar_p$; 0 to retain their old label $\labelvar_p = \labelvar^{'}_p$ or 1 switch to $\alpha$, i.e.~$\labelvar_p = \alpha$.

The Hedgehog term \eqref{eqn:shapepenalty} for a binary \a-exp move could be written as
\begin{equation} \label{eqn:expmoveshapepenalty}
\Hhogs^{\alpha}_\coneangle(\binvars)=\!\!\!\sum_{\!\!\!\!\!\!\!\!\!\!(p,q) \in \shapeedges_{\alpha}(\coneangle)}\!\!\! \shapepairwise_{pq}(\binvar_p,\binvar_q)
+\!\!\!\!\sum_{\labelsetindex \in \labelset \setminus \alpha} \!\underset{\substack{\labelvar^{'}_p = \labelvar^{'}_q=k \\ \\(pq) \in \shapeedges_\labelsetindex (\coneangle)}}{\sum}\!\!\!\!\!\! \! \shapepairwise_{qp}(\binvar_q,\binvar_p)
\end{equation}
where $\binvars=\{\binvar_p \in \{0,1\}\;| \;\forall p \in \pixelset\}$ 
and
\begin{equation}
S_{pq}(\binvar_p,\binvar_q) = \left\{
  \begin{tabular}{cc}
  $\infty$ & if $\binvar_p =1,\; \binvar_q = 0$\\
  0 & \text{otherwise.}
  \end{tabular}
  \right.
\end{equation}
The first term in \eqref{eqn:expmoveshapepenalty} guarantees that the resulting labeling respects label $\alpha$ hedgehog constraints. In addition, the second term guarantees that the hedgehog constraints satisfied by the current labeling $\labelvars^{'}$ for all labels in $\labelset\setminus\alpha$ are not violated by the new labeling $\labelvars$.

According to \cite{Vlad:ECCV02}, any first-order binary function could be exactly optimized if all pairwise terms are submodular. A binary function $g$ of two variables is submodular if $g(0, 0) + g(1, 1) \leq g(1, 0)+g(0, 1)$. Our energy \eqref{eqn:expmoveshapepenalty} is submodular as it could be written as the sum of submodular pairwise binary energies over all possible pairs of $p$ and $q$. Notice that for any given $pq$ pair, $\shapepairwise_{pq}(1,1)=0$ by construction and $\shapepairwise_{pq}(0,0)=0$ as long as the current labeling is a feasible one, i.e.~it does not cut any of the $\infty$-cost arcs. Also, $\shapepairwise_{pq}(1,0)$ and $\shapepairwise_{pq}(0,1)$ are both $\geq 0$ by construction. Therefore, the submodularity condition is satisfied for all pairs of $p$ and $q$.

\begin{figure}[t]
\begin{center}
\begin{tabular}{@{\hspace{-0.5ex}}c@{\hspace{1.0ex}}c}
\includegraphics[width=0.45\linewidth]{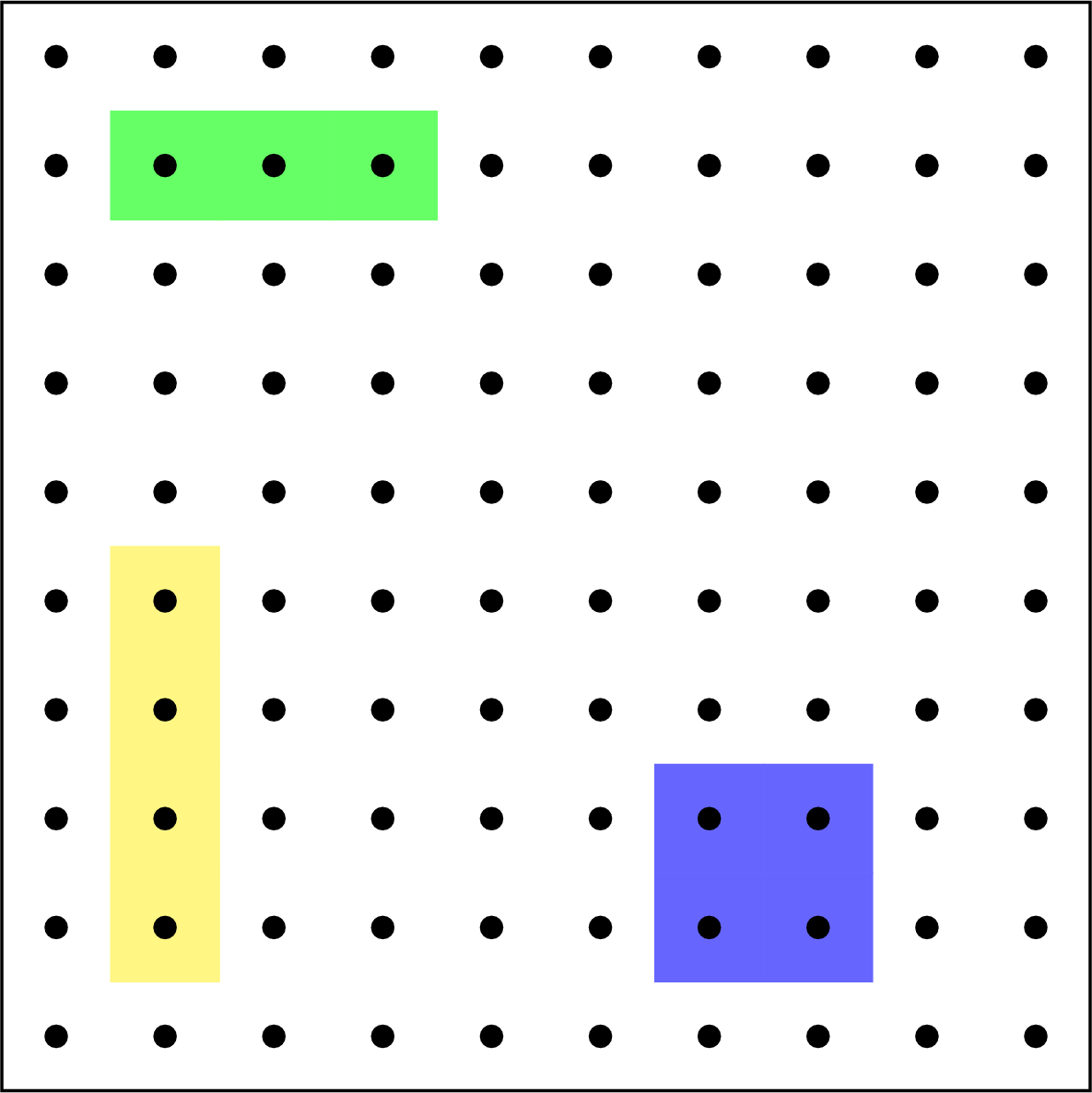}&
\includegraphics[width=0.45\linewidth]{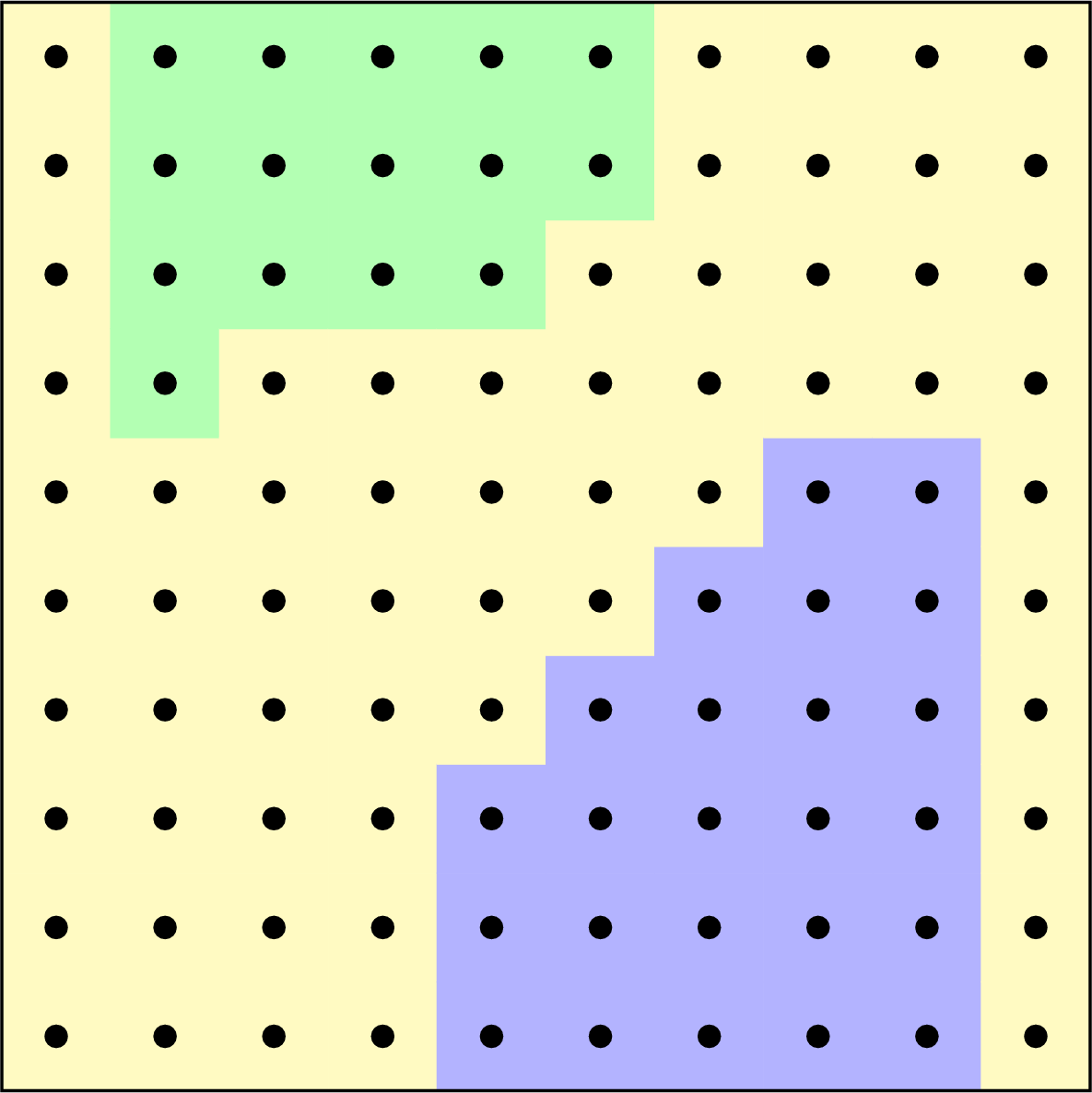}\\
{\small(a) initial Seeds} & {\small(b) current labeling}\\[1ex]
\includegraphics[width=0.45\linewidth]{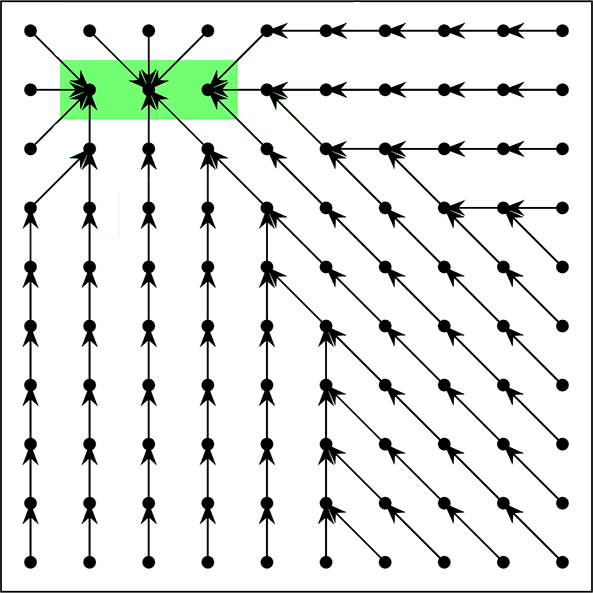}&
\includegraphics[width=0.45\linewidth]{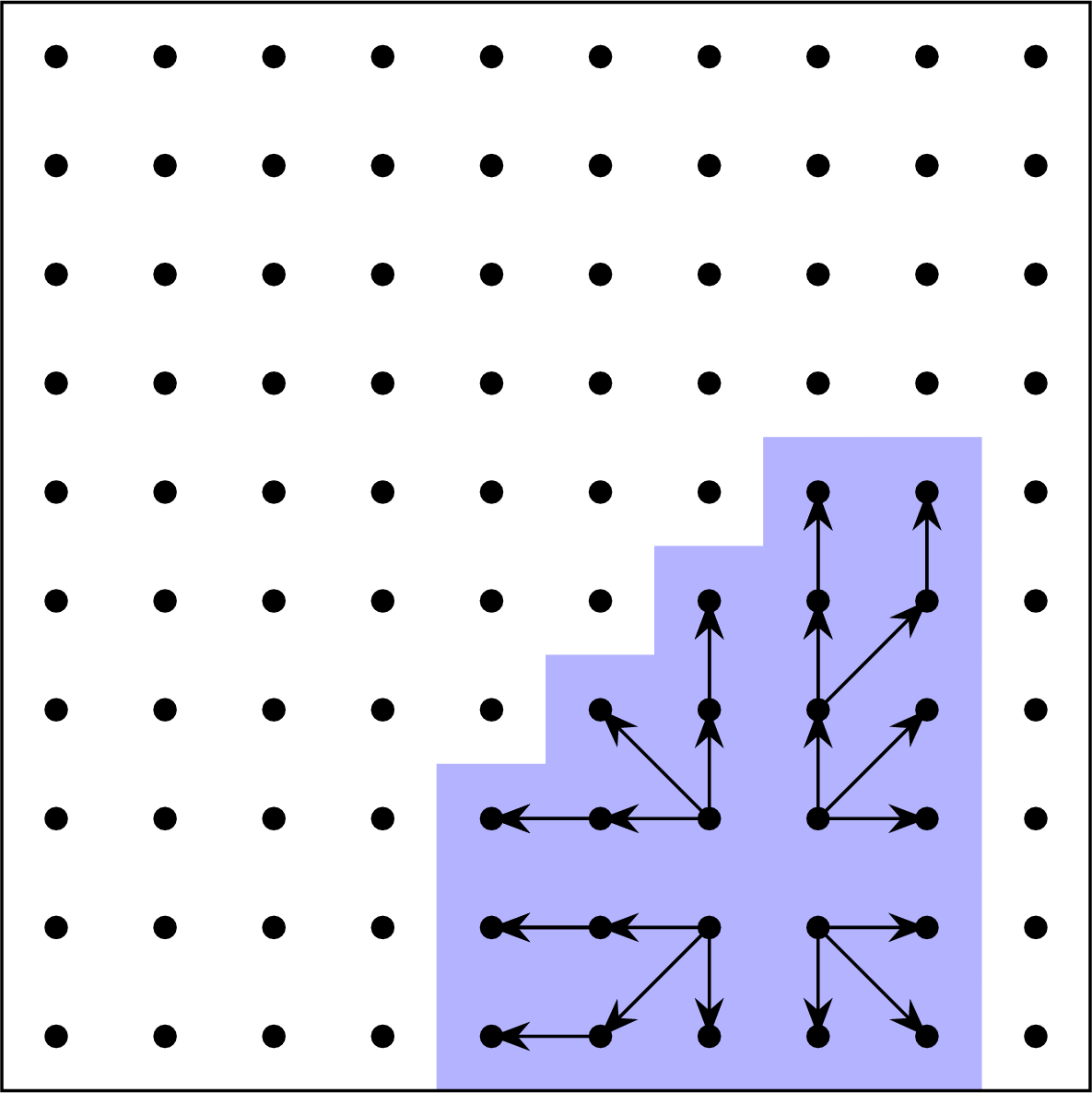}\\
{\small(c) \eqref{eqn:expmoveshapepenalty} first term constraints}&  {\small(d) \eqref{eqn:expmoveshapepenalty} second term constraints}\\[1ex]
\includegraphics[width=0.45\linewidth]{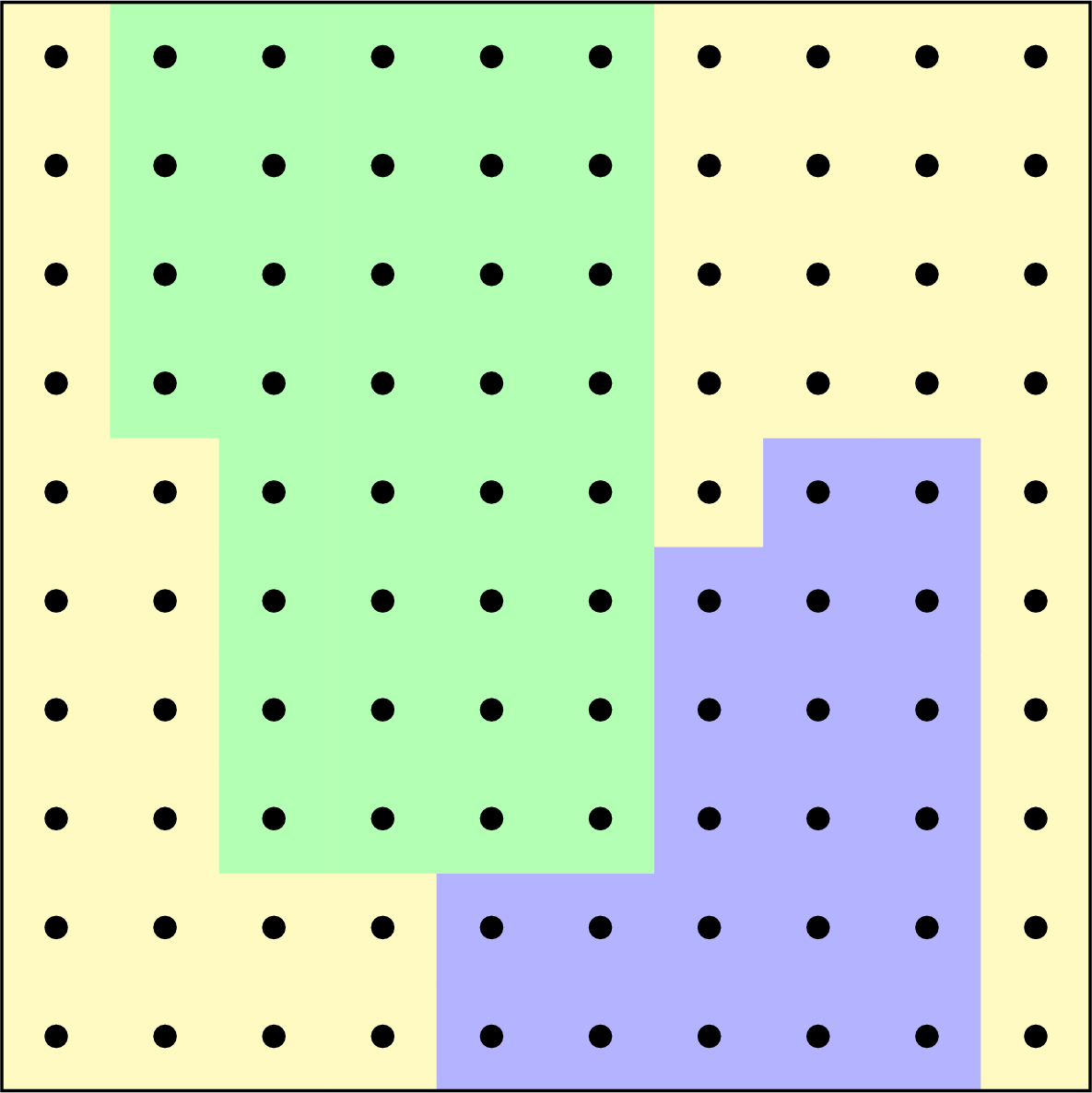}&
\includegraphics[width=0.45\linewidth]{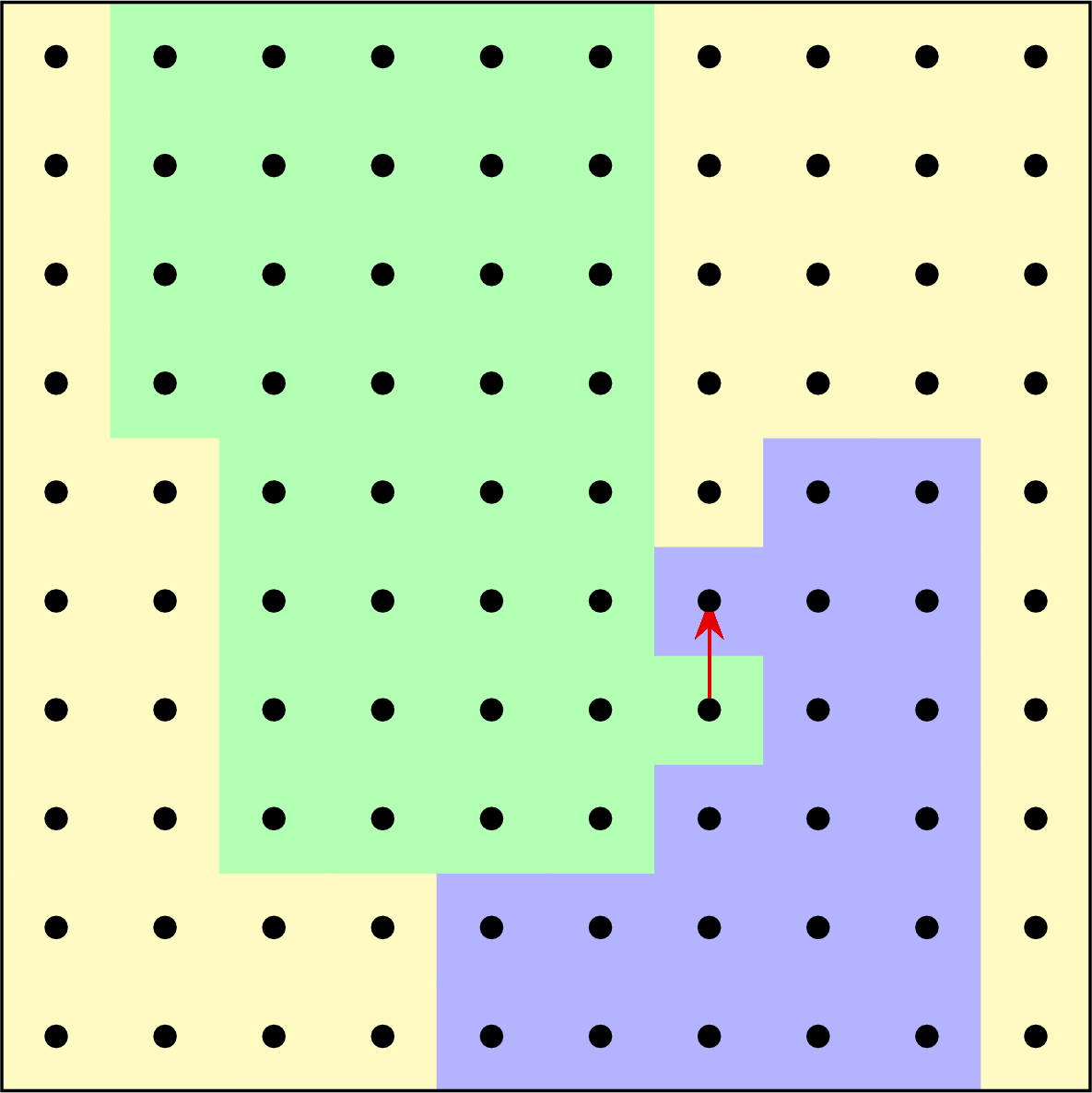}\\
{\small (e) feasible expansion move} &  {\small(f) infeasible expansion move}
\end{tabular}
\end{center}
\caption{{\small Illustration of a feasible and an infeasible expansion move for the green label. (a-b) Initial seeds and current labeling, respectively. (c-d) Hedgehog shape constraints \eqref{eqn:expmoveshapepenalty} enforced by green and purple labels when expanding the green one. (e-f) show a feasible and an infeasible expansion moves, respectively. In (f) severed $\infty$-cost purple shape edge/constraint is shown in red.}}
\label{fig:greenexp}
\end{figure}

Fig.\ref{fig:greenexp} shows an example of an \a-exp move over the green label. We assume the shape constraints only for the green and purple labels. Fig.\ref{fig:greenexp}(a) shows the initial seeds for three different labels while (b) shows the current feasible labeling. Fig.\ref{fig:greenexp}(c-d) show the shape constraints enforced by green and purple labels while expanding the green label. Note, the green shape constraints are enforced all over the image while the purple shape constraints are enforced inside its current labeling support area, as it is not necessarily to enforce it everywhere. Fig.\ref{fig:greenexp}(e) shows a feasible move that respects the green and purple shape constraints while (f) shows an infeasible move that respects only the green shape constraints.

\section{Relation to multi-surface graph cuts} \label{sec:related_graphcuts}

\begin{figure*}[t]
\begin{center}
\begin{tabular}{cc}
{\bf Cartesian discretization approach} & {\bf polar discretization approach} \\[2ex]
\includegraphics[width=0.47\linewidth]{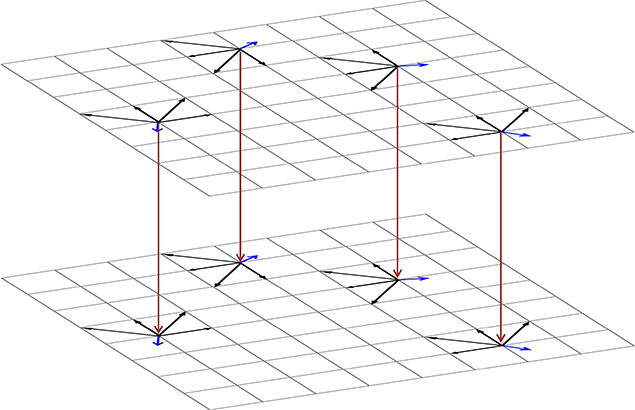}&
\includegraphics[width=0.47\linewidth]{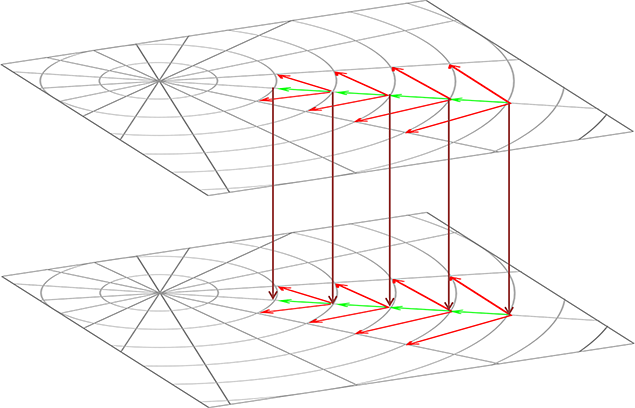}\\
(a) Two identical hedgehogs with inclusion constraint  \cite{DB:ICCV09}  &
(b) Two identical nested {\em stars} as in Logismos \cite{sonka:PAMI06} \\[1ex]
\includegraphics[width=0.47\linewidth]{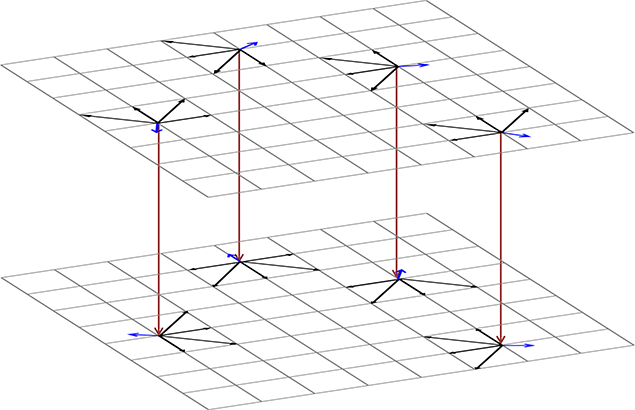}&
\includegraphics[width=0.47\linewidth]{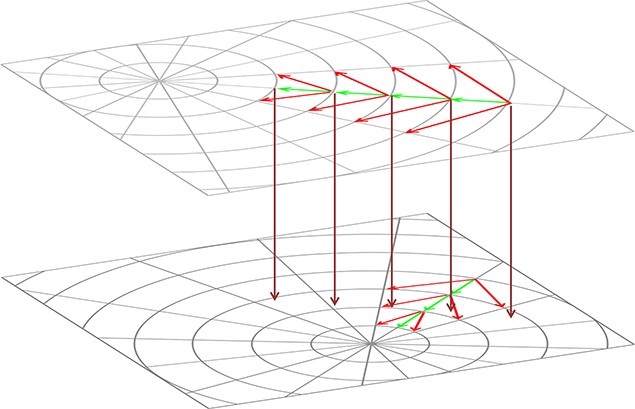}\\
(c) Two distinct hedgehogs with inclusion constraint \cite{DB:ICCV09}   &
(d) Extended {\em multi-polar} Logismos with distinct stars \\
\end{tabular}
\end{center}
   \caption{Multi-object graph cut methods \cite{sonka:PAMI06,DB:ICCV09} with similar inter-surface constraints
and shape priors. Each object corresponds to a layer. The shown inter-layer (brown) edges represent
the simplest examples of inter-surface constraints, \ie~ {\em inclusion} with zero margin \cite{DB:ICCV09} in (a,c) and
{\em nestedness} with zero min distance \cite{sonka:PAMI06} in (b,d). While, the standard regularization of each layer
segment in \cite{DB:ICCV09} is its boundary length \cite{BK:ICCV03}, it can be easily complemented or replaced by shape priors
like star \cite{olga:ECCV08}  or our hedgehog. Integrating the same hedgehog shape (black edges, as in Fig.\ref{fig:twocones})
into both layers (a) creates a Cartesian analogue of Logismos \cite{sonka:PAMI06} (b) based on a vector field (blue) instead of
non-overlapping rays. Using distinct hedgehog shapes (c) is analogous  to an extended multi-polar Logismos (d), which
has some technical issues discussed in Sec.\ref{sec:multipolar}.}
\label{fig:UsVsLogismosMulti}
\end{figure*}

Our work could be related to multi-object segmentation methods \cite{sonka:PAMI06,DB:ICCV09}
combining various forms of boundary regularization and interactions between the surfaces. In particular,
{\em Logismos} \cite{sonka:PAMI06} computes nested segments using polar grid layers
(one per segment) as in Fig.\ref{fig:UsVsLogismosMulti}(b). In general, edges between the layers enforce
inter-surface constraints like minimum and maximum distances between the surfaces along each {\em ray}.
For these constraints to work, the polar grids should be the same at all layers.
Edges within each polar grid enforce regularity of the corresponding segment. Figure
\ref{fig:UsVsLogismosMulti}(b) details the construction.  Red edges penalize
inter-ray surface jumps\footnote{In polar representation, let each segment $i$ correspond to a labeling assigning
distance $L_r^i$ form the pole to the segment boundary along ray $r$. Inter-ray smoothness corresponds to
any convex pairwise potential $U(L_p^i - L_q^i)$ as in Ishikawa \cite{Ishikawa:PAMI03}.
Similarly, inter-layer edges enforcing some min and max distances along each ray $r$  are
a special case of convex potential $U(L_r^i - L_r^j)$.}
and infinity cost green edges enforce a shape prior analogous  to star convexity \cite{olga:ECCV08}.

If considering only one segment, our hedgehog shape prior is closely related to both {\em Logismos} and {\em star convexity}.
The use of Cartezian grid makes our approach closer to methods \cite{KB:ICCV05,olga:ECCV08} already
discussed in Sec.\ref{sec:hhogconst}. Our prior is defined by a vector field, see Fig.\ref{fig:discreteexample}(a),
instead of a polar system of {\em non-overlapping} rays \cite{sonka:PAMI06} requiring considerable care during construction.
Each vector at any of our grid pixels defines a cone of allowed surface normals, see Fig.\ref{fig:twocones},
controlled by width parameter $\theta$. In particular, tighter cones enforce {\em skeleton consistency}.
While our {\em dual cone} of infinity cost edges resembles a combination of green and red edges in each polar layer of Logismos,
our geometrically motivated Cartesian approach uses simpler vector fields generalizing non-overlapping rays and
does not require highly non-uniform polar resampling of images. In fact, our graph construction is technically
different from Logismos, as evident from our discretization details presented in the Appendix.

Also, there are more substantial differences between our multi-hedgehogs method and Logismos.
The latter enforces one star model for all nested shapes since it uses the same polar grids.
In contrast, we do not require nested segments and allow independent shape priors at each segment.
Our current approach does not enforce any geometric inter-segment distances. Thus, it can be seen
as an augmentation of the standard Potts model with independent shape priors for each segment.
However, the following two subsections discuss certain extensions of our multi-hedgehog approach and Logismos
that make them more comparable.

\subsection{Hedgehogs with inter-segment constraints} \label{sec:hedgegogsDB}

If additional geometric inter-segment constraints are needed, our hedgehog shapes could
be easily integrated with the isotropic Cartesian formulations for the {\em inclusion}, {\em minimum margin}, {\em exclusion}
 \cite{DB:ICCV09} and {\em Hausdorf distance} \cite{SB:eccv12}. For example, Fig.\ref{fig:UsVsLogismosMulti}(a) illustrates
a layered graph construction enforcing zero-margin {\em inclusion} for two segments \cite{DB:ICCV09} (brown edges)
combined with the same hedgehog shape prior (black edges, as in Fig.\ref{fig:twocones})
defined by identical vector fields (blue) at two layers. It is also easy to switch to distinct shape priors for each segment
by using different vector fields, see Fig.\ref{fig:UsVsLogismosMulti}(c).

Interestingly, replacing {\em inclusion} by non-submodular {\em exclusion} constraint between the layers
\cite{DB:ICCV09} makes the corresponding model conceptually close to our Potts approach with multi-hedgehog priors.
Thus, our multi-label optimization by $\alpha$-expansion on a single-layer graph in Sec.\ref{sec:expmoves} can be seen as an alternative
to QPBO \cite{QPBO07}, TRWS \cite{TWRS2014}, or other standard approximate optimization methods \cite{kappes2013comparative}
applicable to binary non-submodular multi-layered graphical model in \cite{DB:ICCV09}.
For significant memory savings and potential speed gains, it is possible to reformulate
geometric inter-segment constraints in \cite{DB:ICCV09} as multi-label segmentation potentials that can be addressed with
efficient approximate algorithms on one image-grid layer, \eg $\alpha$-expansion, message passing, or other methods.

\subsection {Multi-polar Logismos with distinct shape priors} \label{sec:multipolar}

It is interesting to consider an option of different polar grids at each layer of Logismos as in
Fig.\ref{fig:UsVsLogismosMulti}(d) that makes it comparable to multi-object segmentation \cite{DB:ICCV09}
with distinct hedgehog shape priors (c) discussed in the previous subsection.
While such {\em multi-polar} extension of Logismos can provide distinct shape priors for the segments,
it raises questions about the inter-layer interactions and their geometric interpretation.
First, there is a minor problem of misalignment between the polar grid nodes.
However, a bigger problem is the misalignment between the rays that calls for a revision of the nestedness
and along-the-ray distance constraints between the surfaces.
If no nestedness is needed, than it is necessary to add non-submodular consistency constraints
between the layers, \ie~ {\em exclusion} \cite{DB:ICCV09}. If nestedness is still required,
then simple inter-surface distance constraints in \cite{sonka:PAMI06} are possible,
but the minimum distance would be enforced along the rays
of the smaller segment layer and the maximum distance would be along the rays of the larger segment.
This discrepancy may be acceptable if the polar systems and the corresponding shape priors are close,
but larger shape differences call for more isotropic definitions of inter-shape distances \cite{DB:ICCV09,SB:eccv12}
that are independent of polar discretization.

\section{Experiments} \label{sec:experiments}

\begin{figure}[t]
\begin{center}
\begin{tabular}{@{\hspace{-0.5ex}}c@{\hspace{1.5ex}}c}
\includegraphics[width=0.50\linewidth]{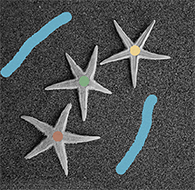}&
\includegraphics[width=0.50\linewidth]{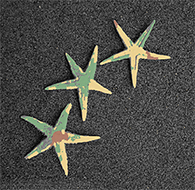}\\
(a) initial seeds & (b) Potts model $\lambda=2$\\
\includegraphics[width=0.50\linewidth]{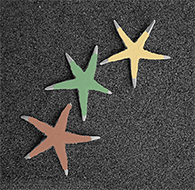}&
\includegraphics[width=0.50\linewidth]{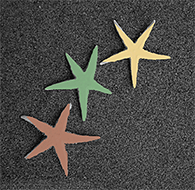}\\
(c) Potts model $\lambda=6$ & (d) Hedgehogs + Potts $\lambda=2$\\
\end{tabular}
\end{center}
   \caption{Three hedgehogs one for each star. (a) shows user scribbles. (c-b) and (d) show Potts model results for different $\lambda$ and our results, respectively. (d) shows that enforcing hedgehogs shape priors (our method) eliminated over-segmented solutions as the one in (b) which is typical for small $\lambda$.
   (c) shows Potts model result for a larger $\lambda$ at which the stars were not over-segmented, notice star tips were wrongly segmented due to the increase in shrinking bias.}
\label{fig:stars}
\vspace{-2ex}
\end{figure}

\begin{figure}[t]
\begin{center}
\begin{tabular}{@{\hspace{-0.5ex}}c@{\hspace{1.5ex}}c}
\includegraphics[width=0.50\linewidth]{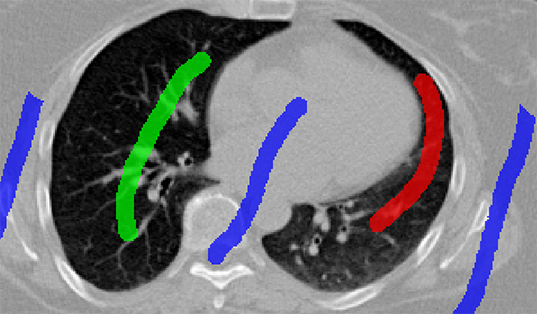}&
\includegraphics[width=0.50\linewidth]{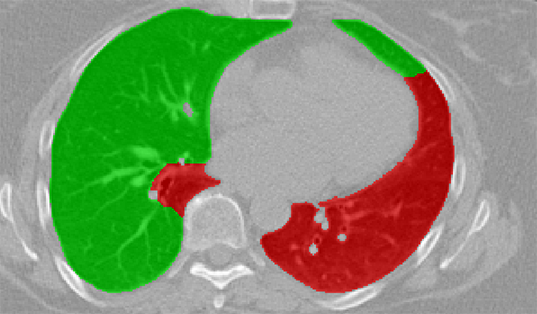}\\
(a) initial seeds & (b) Potts model\\
\includegraphics[width=0.50\linewidth]{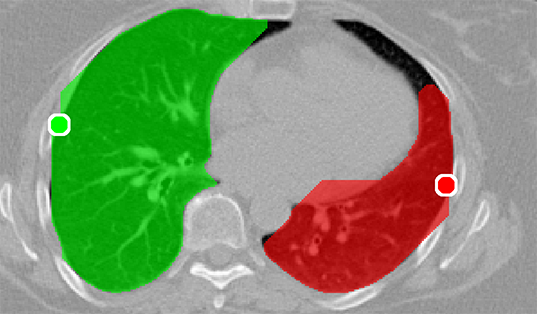}&
\includegraphics[width=0.50\linewidth]{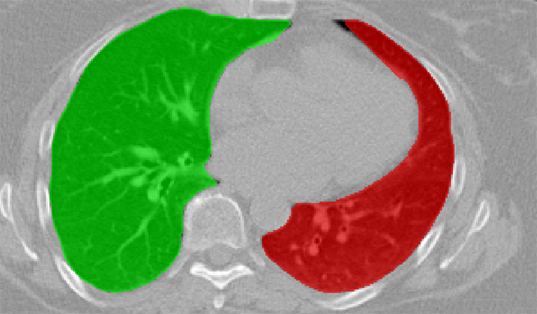}\\
(c) Multi-Star + Potts model & (d) Hedgehogs + Potts model\\
\end{tabular}
\end{center}
   \caption{Two hedgehogs one for each lung. As can be seen in (b) Potts model resulted in segmentation with holes (background inside lungs), and converged to wrong color models.
   Segmentation holes could be eliminated by using multi-star shape priors (c)---star centers are the midpoints of the green and red circles. However, multi-star can never properly segment the right lung as it is not a star-shape. Our method (Hedgehogs + Potts) (d) eliminated holes and properly segmented the lungs by enforcing a more general shape constraint derived from the user scribble.}
\label{fig:lungCrossSection}
\end{figure}

In the following set of experiments we show the benefit of incorporating our {\em Hedgehogs term} \eqref{eqn:shapepenalty} to the well studied Potts model segmentation energy, i.e. {\em data term} + {\em smoothness term}, for multi-object segmentation in 2D and 3D. We will also give an illustrative real life example to show that the hedgehog shape is more general than star-shape \cite{olga:ECCV08}. The results shown in this section for our method were generated using $\theta=\frac{\pi}{4}$ when computing the hedgehog shape constraints, also we did not enforce any shape constraints on the background model. Also, the same smoothness weight $\lambda$ is used when comparing methods unless stated otherwise.

Our optimization framework is similar to \cite{GrabCuts:SIGGRAPH04} where the user marks a set of initial seeds in the form of a scribble for the required labels, e.g.~left kidney, right kidney etc. The seeds for each label were used to fit an initial Gaussian Mixture color model, and to generate its hedgehog shape constraints. Similarly to \cite{PEARL:IJCV12,LabelCosts:IJCV12}, we iteratively optimize our energy \eqref{eqn:OurMainEnergy} (or Potts model) in an EM-style iterative fashion. We alternate between finding a better segmentation and re-estimating the color models using the current segmentation. The framework terminates when it can not decrease the energy anymore.

For the example shown in Fig.\ref{fig:stars}(a), (b-c) show Potts model results for $\lambda=2$ and 6, respectively. It should be noted that $6$ is the smallest smoothness weight that did not result in over-segmentation when using Potts. However, the result in Fig.\ref{fig:stars}(c) is biased towards smaller objects (notice star tips) because by increasing the smoothness weight we are also increasing the shrinking bias. Over-segmented results as the one in Fig.\ref{fig:stars}(b) could be avoided without increasing the shrinking bias, simply by incorporating multi-shape priors. Our method which incorporates Hedgehogs shape priors with Potts model was able to find a better segmentation, see Fig.\ref{fig:stars}(d).

The objective of the example shown in Fig.\ref{fig:lungCrossSection}(a) is to segment left and right lungs, and the background. Potts model result shown in Fig.\ref{fig:lungCrossSection}(b) has holes, i.e.~part of the background appears in the middle of the lungs. Furthermore, Potts model converged to biased color models where the right lung preferred brighter colors while the left preferred darker colors. Similar to the previous example, increasing $\lambda$ for Potts model will increase the shrinking bias and it becomes hard to segment the elongated part of the the right lung. Using multi-star which is a generalization of \cite{olga:ECCV08} to multi-object segmentation is not enough because the right lung is not a star-shape. To be specific, there is no point inside the right lung that could act as a center of a star-shape that would include it. Fig.\ref{fig:lungCrossSection}(d) shows the result for our method, where user scribbles were used to enforce shape constraints compared to using a single pixel per label \cite{olga:CVPR10}.

\begin{figure*}
\vspace{-2.5em}
\begin{tabular}{@{\hspace{-4.2ex}}c@{\hspace{0.5ex}}c@{\hspace{0.5ex}}c@{\hspace{0.5ex}}c@{\hspace{0.5ex}}c@{\hspace{0.5ex}}c}
& & {\large Subject 1} & {\large Subject 2}& {\large Subject 3}\\[-5ex]
&\multirow{5}{*}{{\rdelim\{{36}{2.5mm}}} & & & \\[2.5ex]
\multirow{3}{*}{\begin{sideways}Our method (Hedgehogs Shapes + Potts)\phantom{\quad\quad}\end{sideways}}&&
\includegraphics[width=0.33\linewidth]{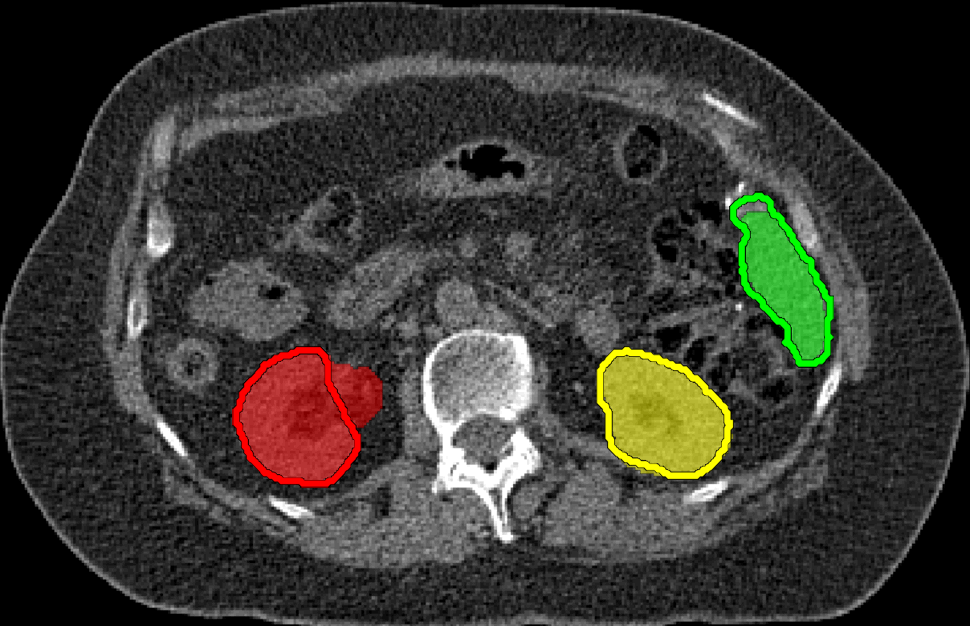}    &
\includegraphics[width=0.33\linewidth]{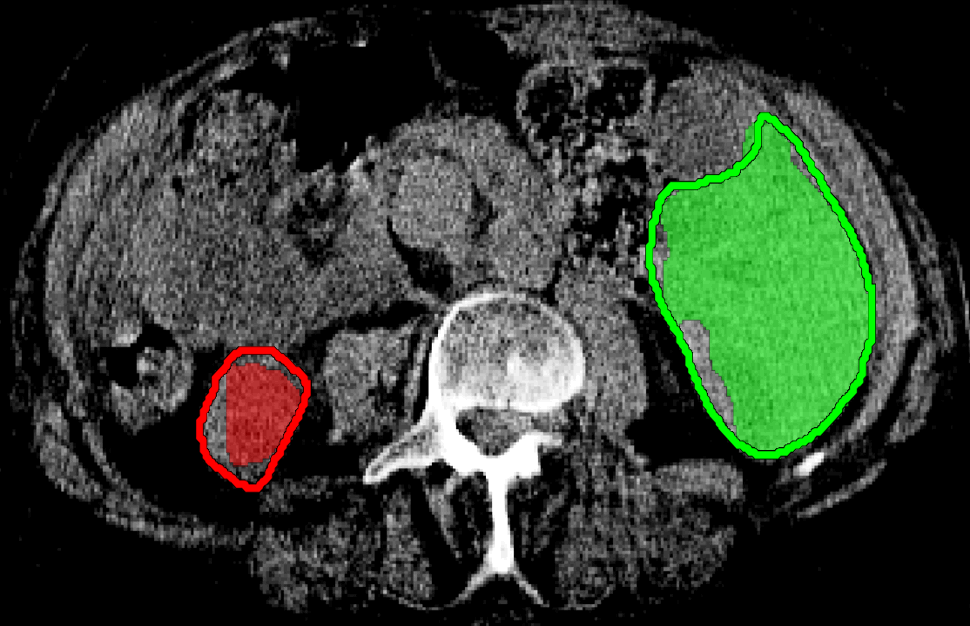}  &
\includegraphics[width=0.33\linewidth]{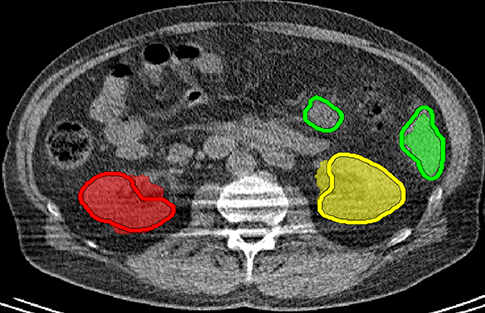} &\\
&&
\includegraphics[width=0.33\linewidth]{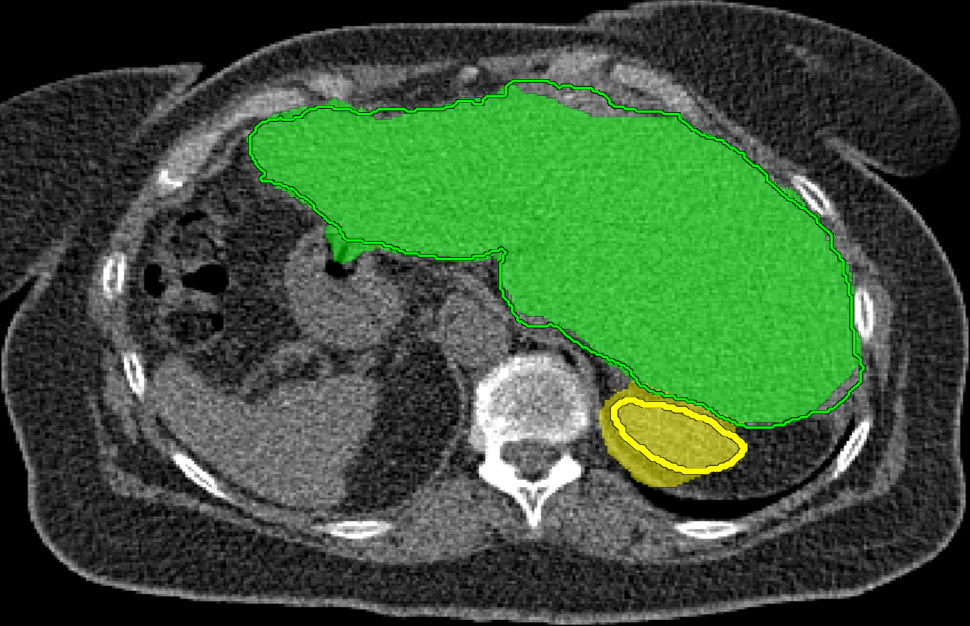}    &
\includegraphics[width=0.33\linewidth]{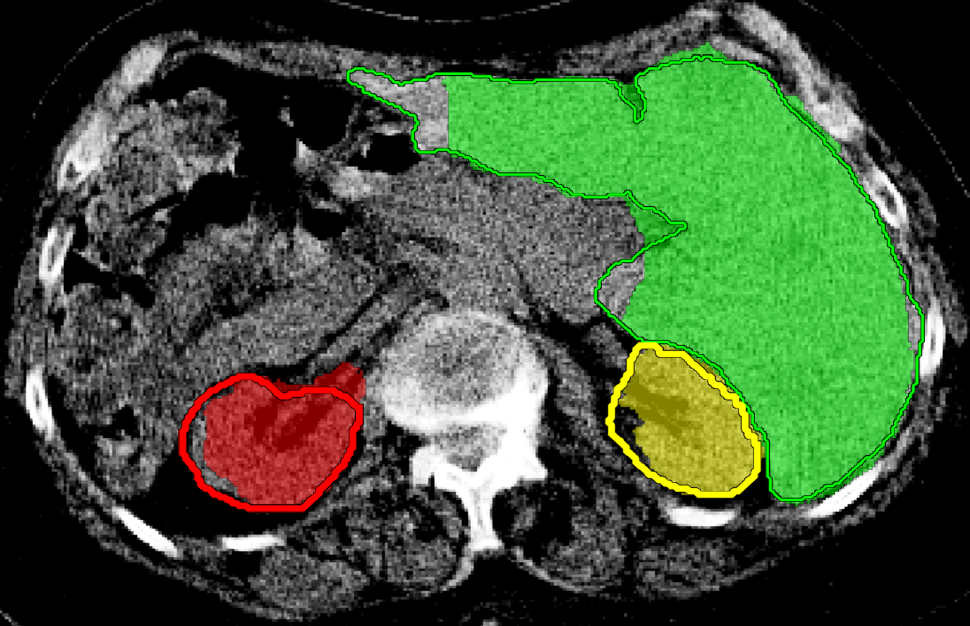}  &
\includegraphics[width=0.33\linewidth]{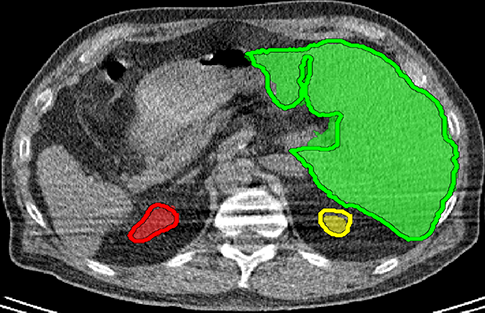} & \\
&&
\includegraphics[width=0.33\linewidth]{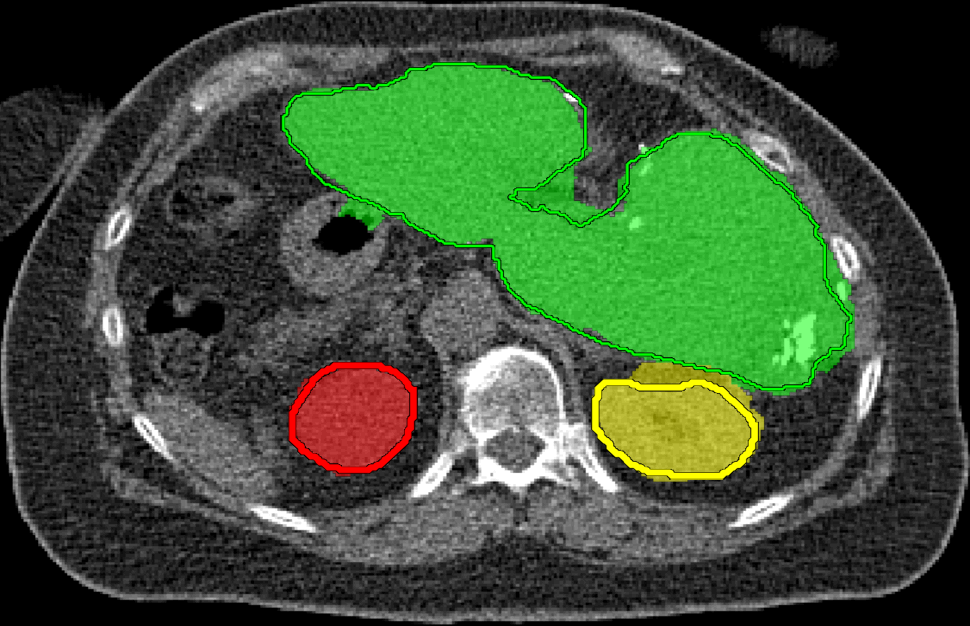}    &
\includegraphics[width=0.33\linewidth]{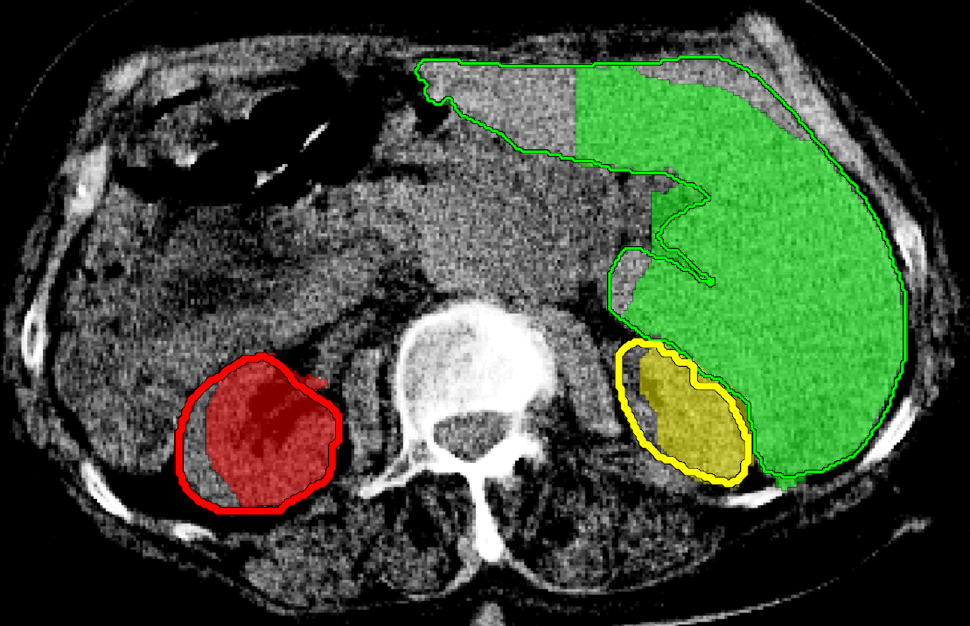} &
\includegraphics[width=0.33\linewidth]{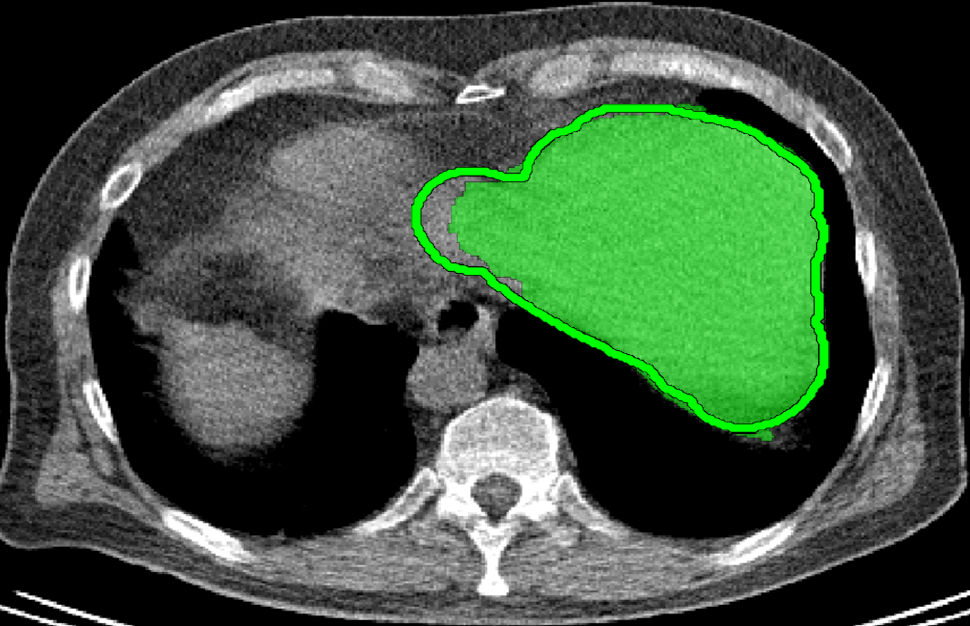} \\ [-2.5ex]
&&&&& \multirow{2}{*}{\rdelim\}{19}{10pt} \begin{rotate}{-90}\phantom{mmmmmmmmmmm}Same Slice \end{rotate} } \\
&&
\includegraphics[width=0.33\linewidth]{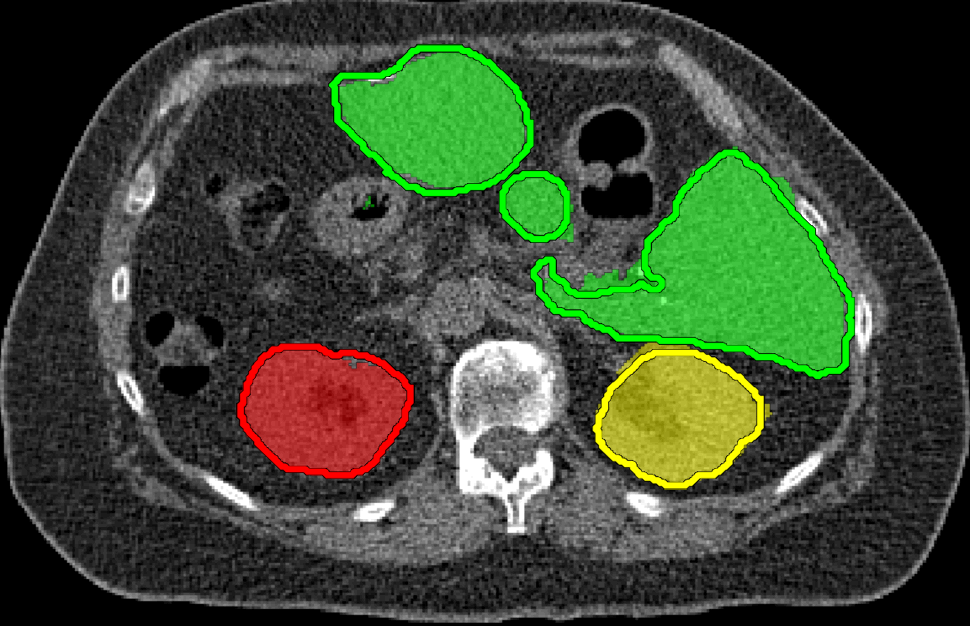}&
\includegraphics[width=0.33\linewidth]{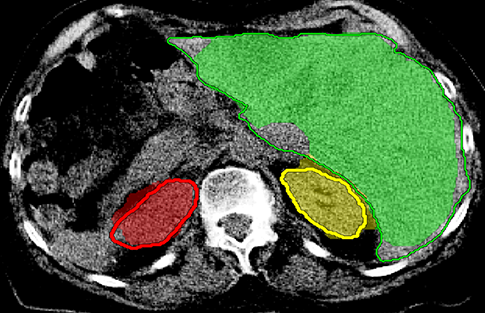}&
\includegraphics[width=0.33\linewidth]{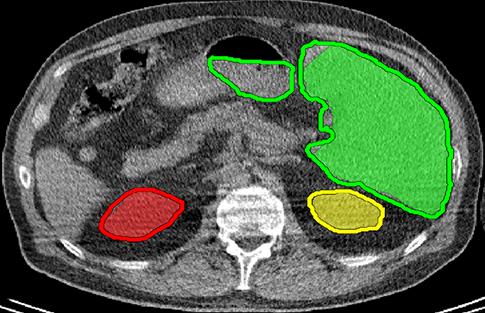}& \\[1.5ex]
\begin{sideways}\phantom{mmmmm}Potts\end{sideways}&&
\includegraphics[width=0.33\linewidth]{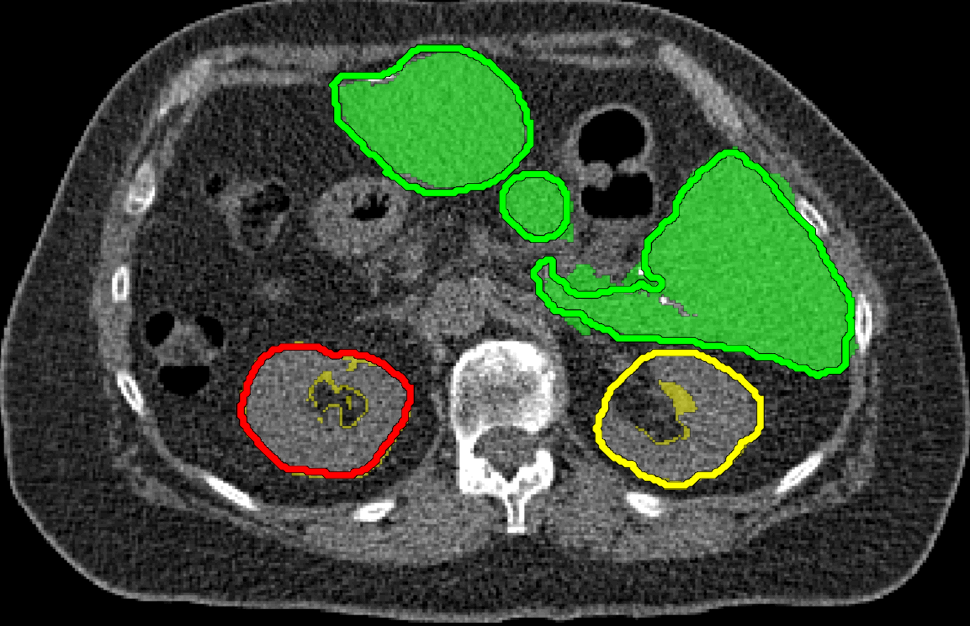} &
\includegraphics[width=0.33\linewidth]{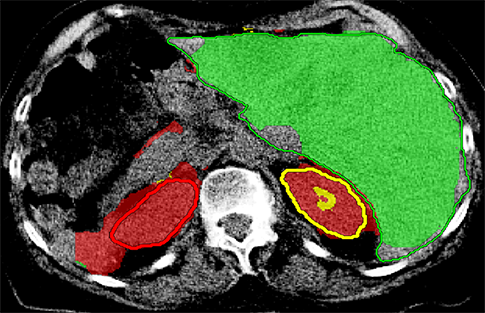} &
\includegraphics[width=0.33\linewidth]{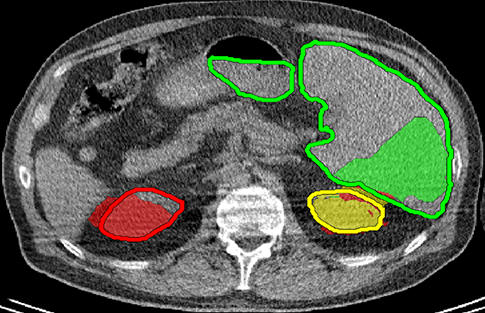} &
\end{tabular}
\caption{Three hedgehogs for liver and two kidneys, the colored contours represent liver and kidneys ground truth. Each column shows the result of a different test subject. The first four rows show our results, each row represents a different slice. The last row shows Potts model results. Also, the last two rows show results of the same slice for our method and Potts model, respectively. Our method (Hedgehogs+Potts) out performed Potts, results show that enforcing shape constraints avoids/forbids some undesirable segmentations, e.g.~for subject 2 Potts segmentation shows that the left kidney surrounds the right kidney, and for subject 3 it shows that part of the left kidney is between the right kidney and liver. In addition, for subjects 1 and 2 Potts model did not properly separate the kidneys from the background.}
\label{fig:liverandkidneys}
\end{figure*}

We applied our method on PET-CT scans of three different subjects to segment their liver, left kidney, right kidney and the background. Although we applied our method and Potts model on the 3D volumes we only show the results on a few representative slices from each volume in Fig.\ref{fig:liverandkidneys}. Also, the results of different methods for each subject were computed using the same smoothness.
We can see from the last two rows which compare our method to Potts, using Hedgehogs constraints enabled us to avoid geometrically incorrect segmentations, e.g.~one liver inside the other (last-row middle), or parts of left kidney is between the right kidney and liver (last-row right).
Furthermore, for test subjects 1 and 2 the kidneys and background were poorly segmented by Potts model, e.g.~most of the kidneys were segmented as background for test subject 1. Potts poor performance is due to the large overlap between the kidneys and background color models. This overlap resulted in an in-discriminative data term for Potts to properly separate them. This issue becomes worse in iterative frameworks where color models are re-estimated based on current segmentation. To be specific, if at any iteration Potts model resulted in a bad segmentation then re-estimating the color models will bias them towards the bad segmentation and subsequent iterations worsen the results. Comparing our results for subjects 1 and 2 to Potts model shows that our method is less prone to the aforementioned issue as we forbid undesirable segmentations, i.e.~those that do not respect shape constraints.

For quantitative comparison, Table \ref{tbl:liverKindeysTable} lists for each organ of a subject the $\mathbf{F}_1$ Score, Precession and Recall measures of our method and Potts model where \mbox{$\mathbf{F}_1=2*\frac{Precession *Recall}{Precession+Recall}.$} For the kidneys, our method clearly out performed Potts model, e.g.~note Potts model poor precision/recall for subjects 1 and 2.
For the liver, both methods performed comparably.

\definecolor{liver}{RGB}{0,255,0}
\definecolor{k1}{RGB}{255,255,0}
\definecolor{k2}{RGB}{255,0,0}

\begin{table}[h]
\begin{center}
\begin{tabular}{c|c|c|c|c|c|c|}
\cline{2-7}
& \multicolumn{2}{c}{Subject 1} & \multicolumn{2}{|c}{Subject 2} & \multicolumn{2}{|c|}{Subject 3}\\
\cline{2-7}
 &Ours & Potts & Ours & Potts & Ours & Potts\\
\cline{2-7}
\multicolumn{7}{c}{\vspace{-2ex}}\\
\cline{2-7}
&\multicolumn{6}{c|}{Right Kidney \crule[k1]{0.2cm}{0.2cm}}\\
\hline
\multicolumn{1}{|c|}{$\mathbf{F}_1$ score} &$\mathbf {0.85}$ & $0.05$ & $ \mathbf {0.69}$ & $0.11 $&  $\mathbf {0.92} $ & $0.85 $\\
\hline
\multicolumn{1}{|c|}{Prec.} &$0.77$ & $0.16$ & $0.58$ & $0.13$&  $0.93$ & $0.85$\\
\hline
\multicolumn{1}{|c|}{Recall} &$0.96$ & $0.03$ & $0.84$ & $0.10$&  $0.91$ & $0.87$\\
\hline
\multicolumn{7}{c}{\vspace{-2ex}}\\
\cline{2-7}
&\multicolumn{6}{c|}{Left Kidney \crule[k2]{0.2cm}{0.2cm}}\\
\hline
\multicolumn{1}{|c|}{$\mathbf{F}_1$ score} &$\mathbf {0.96}$ & $0.08$ & $\mathbf {0.81}$ & $0.48 $&  $\mathbf {0.93}$ & $0.84 $\\
\hline
\multicolumn{1}{|c|}{Prec.}& $0.90$ & $0.97$&$0.85$  & 0.34&  $0.95$& $0.76$\\
\hline
\multicolumn{1}{|c|}{Recall} & $0.95$ & $0.04$& $0.78$ & $0.80$&  $0.91$& $0.93$\\
\hline
\multicolumn{7}{c}{\vspace{-2ex}}\\
\cline{2-7}
&\multicolumn{6}{c|}{Liver \crule[liver]{0.2cm}{0.2cm}}\\
\hline
\multicolumn{1}{|c|}{$\mathbf{F}_1$ score} &$0.92$ & $\mathbf {0.93}$ & $ 0.90$ & $\mathbf {0.91} $&  $\mathbf {0.92}$ & $0.84 $\\
\hline
\multicolumn{1}{|c|}{Prec.}& $0.92$ & $0.93$&  $0.97$& 0.96&  $0.97$& $0.96$\\
\hline
\multicolumn{1}{|c|}{Recall}& $0.92$ & $0.93$&$0.84$  &$0.87$ &  $0.88$& $0.74$ \\
\hline
\end{tabular}
\vspace{2ex}
\caption{The table lists the $\mathbf{F}_1$ score, precision and recall measures for each method, individual organ and subject---the closer these values are to 1 the more accurate the segmentation is. For the kidneys where most of the color model overlap occurs, our method was a clear winner. For the liver which has a bigger volume and a more distinct color model compared to the kidneys/background, the two methods performed comparably.  }
\vspace{-6ex}
\label{tbl:liverKindeysTable}
\end{center}
\end{table}

\section{Conclusion}
We proposed a novel interactive multi-object segmentation method where objects are restricted by hedgehog
shapes.
The hedgehog shape constraints of an object limits its set of possible segmentations by restricting the segmentation's allowed surface normal orientations.
Hedgehog shape constraints could be derived from some vector field, e.g.~the gradient of a user scribble distance transform.
In addition, we showed how to modify $\alpha$-expansion moves to optimize our multi-labeling problem with hedgehog constraints.
We also proved submodularity of the modified binary expansion moves.
Furthermore, we applied our multi-labeling segmentation with hedgehog shapes on 2D images and 3D medical volumes.
Our experiments show the significant improvement in segmentation accuracy when using our method over Potts model.
Specially in medical data where our method outperformed Potts model in separating multiple organs with similar appearances and weak edges.
\appendix
\vspace{-0.5ex}
\section*{Appendix: Discretization Issues}
\vspace{-0.5ex}
There are some challenges/drawbacks due to the discretization of hedgehog constraints. For example, the number of representable surface orientations depends on the chosen neighborhood system $\neighbset$ which could be remedied by using larger neighborhood systems. Also it is possible for a polar cone ${\hat C}_{\theta}(p)$ to be under represented by $\shapeedges(\theta)$ if it happens that no edges lie in it which could result in a segmentation surface with folds. Furthermore, in cases where the vector field changes relatively fast w.r.t.~the image resolution it is possible for neighboring pixel's hedgehog constraints to conflict.
\begin{figure}
\begin{center}
\begin{tabular}{@{\hspace{-0.5ex}}c@{\hspace{1.5ex}}c}
\includegraphics[width=0.50\linewidth]{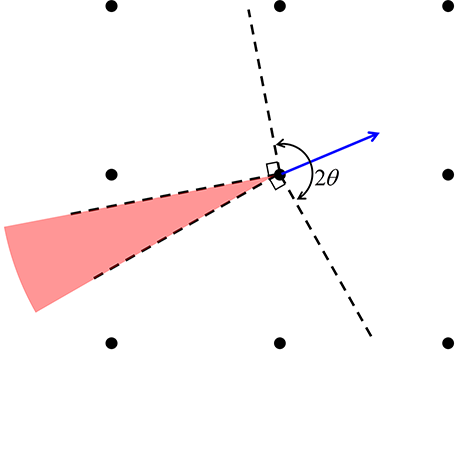}&
\includegraphics[width=0.50\linewidth]{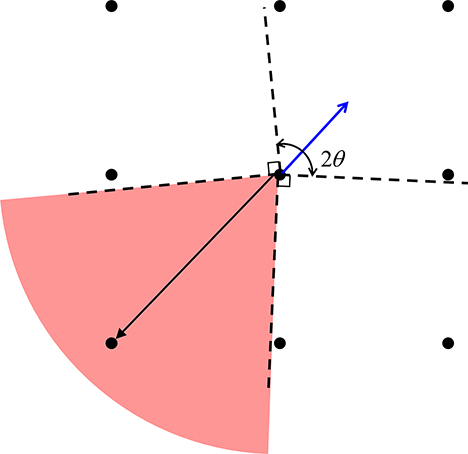}\\
\small{(a) Empty Cone} & \small{(b) Under-represented cone}\\
\\
\includegraphics[width=0.50\linewidth]{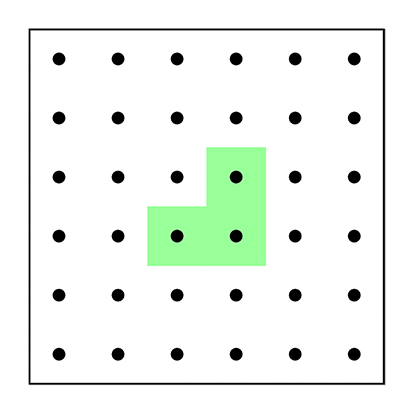}&
\includegraphics[width=0.50\linewidth]{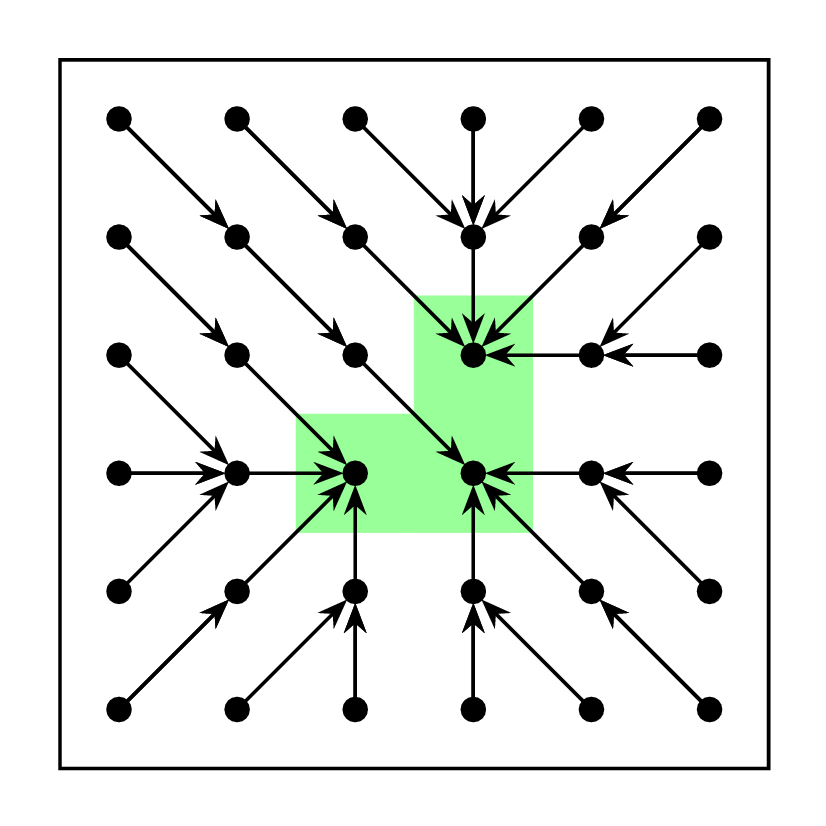}\\
\small{(c) $\shapeedges(\frac{\pi}{2})=\phi$} & (d) \small{Alternative $\shapeedges(\frac{\pi}{2})$}\\
\end{tabular}
\end{center}
\caption{(a-b) illustrate the two cases where a polar cone (shown in red) is under represented by $\shapeedges(\theta)$ (shown as black directed edges). (c) shows $\shapeedges(\frac{\pi}{2})$ of the surface normals/gradient shown in Fig.~\ref{fig:discreteexample}(a). Note that $\shapeedges(\frac{\pi}{2})$ is empty because none of surface normals at any pixel align with its 8-neighbourhood edges. (d) shows the alterative $\shapeedges(\frac{\pi}{2})$ where the nearest neighbourhood edge to the empty cone is added as a shape constraint.}
\vspace{-2ex}
\label{fig:emptyconeissue}
\end{figure}

\noindent\paragraph{Cone under-representation:} a pixel's polar cone ${\hat C}_{\theta}(p)$  is under represented by $\shapeedges(\theta)$ in two case: (a) {\em ``empty cone ''} when there are no neighbor edges consistent with the polar cone as shown in Fig.~\ref{fig:emptyconeissue}(a), and (b) when there is a large part of the cone unaccounted for, see Fig.~\ref{fig:emptyconeissue}(b) where the big cone is accounted for by only one edge. Based on our practical experience only ignoring the former case is of significant consequences while ignoring later does not adversary affect the results.

Empty cone issue could be alleviated by increasing the neighborhood size. However, this is not practical because for $\theta=\frac{\pi}{2}$ the neighborhood edge has to perfectly align with the surface normal, see Fig.~\ref{fig:emptyconeissue}(c). Alternatively, we propose adding to $\shapeedges(\theta)$ the nearest neighborhood edge to the empty cone as shown in Fig.\ref{fig:emptyconeissue}(d).
\begin{figure*}
\begin{center}
\begin{tabular}{ccc}
\includegraphics[width=0.3\linewidth]{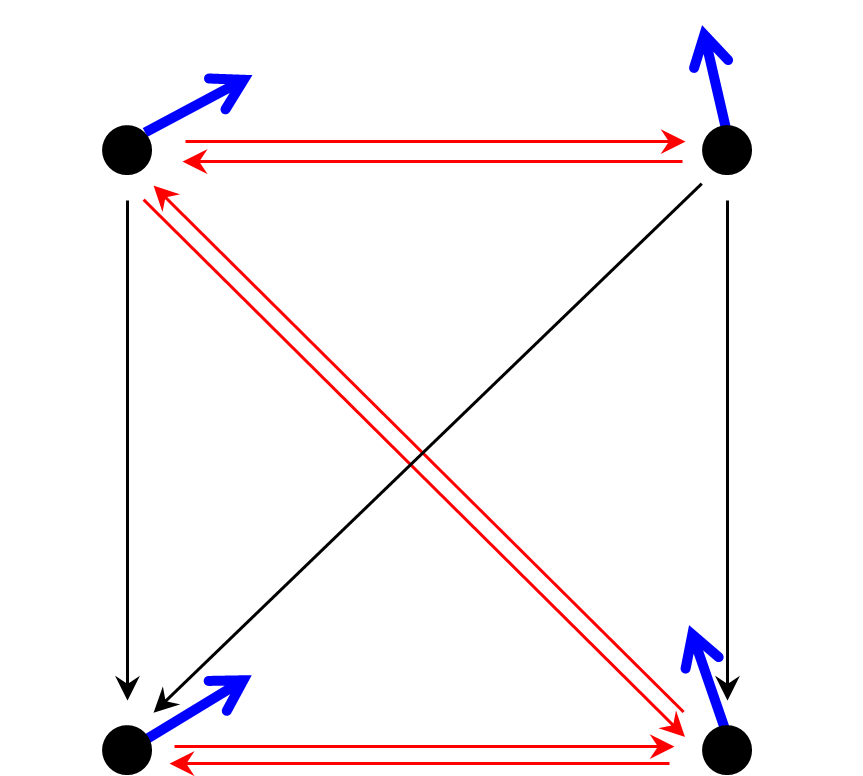} &
\includegraphics[width=0.3\linewidth]{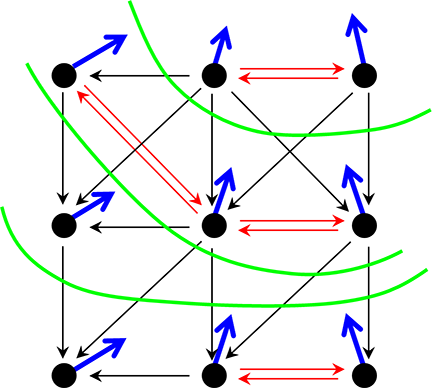} &
\includegraphics[width=0.3\linewidth]{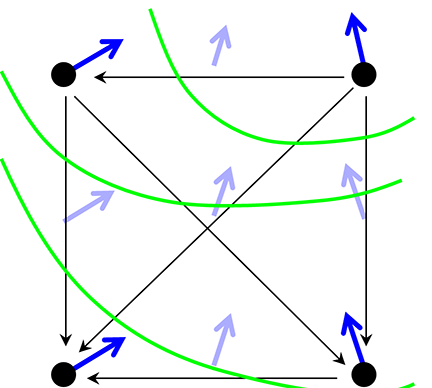} \\
(a) over-constrained $\shapeedges(0)$ & (b) $\shapeedges(0)$ with higher resolution & (c) $\shapeedges(0)$ after edge pruning
\end{tabular}
\end{center}
\caption{Permissible segmentations shown in green, $\shapeedges(0)$ shown in black and red, and vector field shown in blue. (a) shows $\shapeedges(0)$ where over-constraining occurs due to the rapid change in orientation. (b) shows how increasing the solution could solve the over-constraining issue. (c) shows the case where edges in $\shapeedges(0)$ were eliminated when they were inconsistent with the interpolated vector field (shown in light blue).}
\label{fig:solutionABC}
\vspace{1ex}
\end{figure*}
\begin{figure*}
\begin{center}
\begin{tabular}{ccc}
\includegraphics[width=0.3\linewidth]{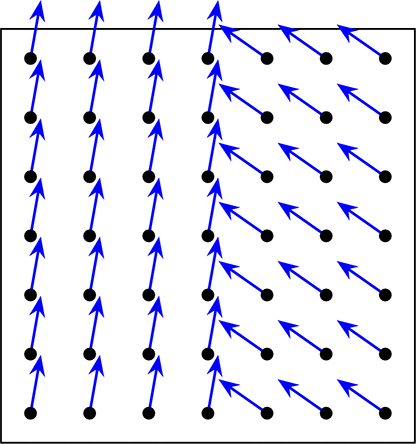} &
\includegraphics[width=0.3\linewidth]{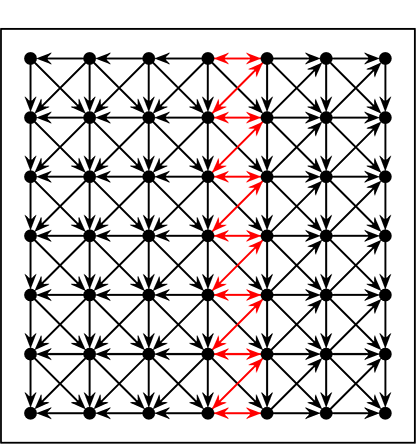} &
\includegraphics[width=0.3\linewidth]{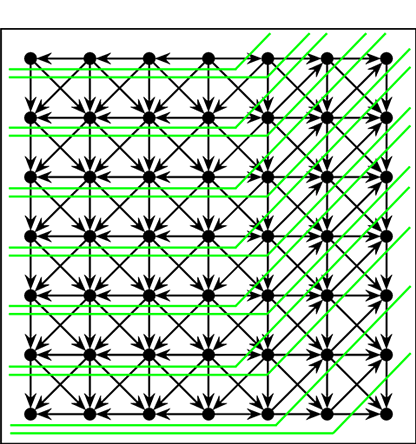} \\
(a) vector field with rapid changes & (b) $\shapeedges(0)$ & (c) $\shapeedges(0)$ after pruning
\end{tabular}
\end{center}
\caption{(a) vector field. (b) $\shapeedges(0)$ of (a) where over-constraining occurs (shown in red). In (b) only the trivial segmentations are allowed due to over-constraining. (c) shows $\shapeedges(0)$ after pruning and the non-trivial segmentations (shown in green) allowed by that construction.}
\label{fig:ironrodexample}
\end{figure*}

\noindent\paragraph{Fast changing vector field:} hedgehog prior \eqref{eq:e_theta} enforces the shape constraints at every pixel independently. When the vector field orientation changes rapidly between neighboring pixels the resulting shape constraints could become contradictory leading to over-constraining. As can be seen in Fig.~\ref{fig:solutionABC}(a) the contradictory shape constraints resulted in a construction where no surface could path between the four neighboring pixels, i.e.~all of them will either be labeled foreground or background.

One possible way to overcome the fast changing vector fields is to increase the image resolution via sub-sampling. As you can see in Fig.~\ref{fig:solutionABC}(b) doubling the resolution alleviated the aforementioned issue. However, there is no simple answer to at which resolution there will be no over-constraining, as it will depend on $\theta$, $\neighbset$ and the vector field. Also, increasing the image resolution is not a practical solution as it adversely affects the running time.

Alternatively one can try an resolve contradicting constraints by pruning $\shapeedges(\theta)$. In this case, we interpolate the vector field's orientation for every neighboring pixels and eliminate  their edge constraint(s) if it were not consistent with the interpolated orientation, as shown in Fig.~\ref{fig:solutionABC}(c). Fig\ref{fig:ironrodexample} shows a synthetic example of fast chancing vector field orientations and how the edge constraints pruning alleviates the over-constraining problem.

{\small
\bibliographystyle{ieee}
\bibliography{hedgehogs_arxiv15}

\begin{thebibliography}{10}\itemsep=-1pt

\bibitem{ConvexMultiRegionICCV2011}
S.~Andrews, C.~McIntosh, and G.~Hamarneh.
\newblock Convex multi-region probabilistic segmentation with shape prior in
  the isometric log-ratio transformation space.
\newblock In {\em Computer Vision (ICCV), 2011 IEEE International Conference
  on}, pages 2096--2103. IEEE, 2011.

\bibitem{BJ:01}
Y.~Boykov and M.-P. Jolly.
\newblock {\em Interactive graph cuts} for optimal boundary \& region
  segmentation of objects in {N-D} images.
\newblock In {\em ICCV}, volume~I, pages 105--112, July 2001.

\bibitem{BK:ICCV03}
Y.~Boykov and V.~Kolmogorov.
\newblock Computing geodesics and minimal surfaces via graph cuts.
\newblock In {\em International Conference on Computer Vision}, volume~I, pages
  26--33, 2003.

\bibitem{BVZ:ICCV99}
Y.~Boykov, O.~Veksler, and R.~Zabih.
\newblock Fast approximate energy minimization via graph cuts.
\newblock In {\em International Conference on Computer Vision}, volume~I, pages
  377--384, 1999.

\bibitem{BVZ:PAMI01}
Y.~Boykov, O.~Veksler, and R.~Zabih.
\newblock Fast approximate energy minimization via graph cuts.
\newblock {\em IEEE transactions on Pattern Analysis and Machine Intelligence},
  23(11):1222--1239, November 2001.

\bibitem{cremers2008shape}
D.~Cremers, F.~R. Schmidt, and F.~Barthel.
\newblock Shape priors in variational image segmentation: Convexity, lipschitz
  continuity and globally optimal solutions.
\newblock In {\em Computer Vision and Pattern Recognition, 2008. CVPR 2008.
  IEEE Conference on}, pages 1--6. IEEE, 2008.

\bibitem{DB:ICCV09}
A.~Delong and Y.~Boykov.
\newblock {Globally Optimal Segmentation of Multi-Region Objects}.
\newblock In {\em International Conference on Computer Vision (ICCV)}, 2009.

\bibitem{LabelCosts:IJCV12}
A.~Delong, A.~Osokin, H.~Isack, and Y.~Boykov.
\newblock {Fast Approximate Energy Minization with Label Costs}.
\newblock {\em International Journal of Computer Vision (IJCV)}, 96(1):1--27,
  January 2012.

\bibitem{olga:CVPR10}
P.~F. Felzenszwalb and O.~Veksler.
\newblock Tiered scene labeling with dynamic programming.
\newblock In {\em Computer Vision and Pattern Recognition (CVPR)}, 2010.

\bibitem{GeodesicStarCVPR10}
V.~Gulshan, C.~Rother, A.~Criminisi, A.~Blake, and A.~Zisserman.
\newblock Geodesic star convexity for interactive image segmentation.
\newblock In {\em Proceedings of the IEEE Conference on Computer Vision and
  Pattern Recognition}, 2010.

\bibitem{PEARL:IJCV12}
H.~N. Isack and Y.~Boykov.
\newblock {Energy-based Geometric Multi-Model Fitting}.
\newblock {\em International Journal of Computer Vision (IJCV)},
  97(2):123--147, April 2012.

\bibitem{Ishikawa:PAMI03}
H.~Ishikawa.
\newblock Exact optimization for {Markov Random Fields} with convex priors.
\newblock {\em IEEE transactions on Pattern Analysis and Machine Intelligence},
  25(10):1333--1336, 2003.

\bibitem{kappes2013comparative}
J.~Kappes, B.~Andres, F.~Hamprecht, C.~Schnorr, S.~Nowozin, D.~Batra, S.~Kim,
  B.~Kausler, J.~Lellmann, N.~Komodakis, et~al.
\newblock A comparative study of modern inference techniques for discrete
  energy minimization problems.
\newblock In {\em Computer Vision and Pattern Recognition (CVPR)}, pages
  1328--1335, 2013.

\bibitem{KB:ICCV05}
V.~Kolmogorov and Y.~Boykov.
\newblock What metrics can be approximated by geo-cuts, or global optimization
  of length/area and flux.
\newblock In {\em International Conference on Computer Vision}, October 2005.

\bibitem{Vlad:ECCV02}
V.~Kolmogorov and R.~Zabih.
\newblock What energy functions can be minimized via graph cuts.
\newblock In {\em 7th European Conference on Computer Vision}, volume III of
  {\em LNCS 2352}, pages 65--81, Copenhagen, Denmark, May 2002.
  {Springer-Verlag}.

\bibitem{sonka:PAMI06}
K.~Li, X.~Wu, D.~Z. Chen, and M.~Sonka.
\newblock Optimal surface segmentation in volumetric images-a graph-theoretic
  approach.
\newblock {\em IEEE transactions on Pattern Analysis and Pattern Recognition
  (PAMI)}, 28(1):119--134, January 2006.

\bibitem{pizer2003deformable}
S.~M. Pizer, P.~T. Fletcher, S.~Joshi, A.~Thall, J.~Z. Chen, Y.~Fridman, D.~S.
  Fritsch, A.~G. Gash, J.~M. Glotzer, M.~R. Jiroutek, et~al.
\newblock Deformable m-reps for 3d medical image segmentation.
\newblock {\em International Journal of Computer Vision}, 55(2-3):85--106,
  2003.

\bibitem{Mreps:IJCV03}
S.~Pizer~{\em et al.}
\newblock {Deformable M-Reps for 3D Medical Image Segmentation}.
\newblock {\em International Journal of Computer Vision (IJCV)},
  55(2-3):85--106, November 2003.

\bibitem{GrabCuts:SIGGRAPH04}
C.~Rother, V.~Kolmogorov, and A.~Blake.
\newblock Grabcut - interactive foreground extraction using iterated graph
  cuts.
\newblock In {\em ACM transactions on Graphics (SIGGRAPH)}, August 2004.

\bibitem{QPBO07}
C.~Rother, V.~Kolmogorov, V.~Lempitsky, and M.~Szummer.
\newblock Optimizing binary mrfs via extended roof duality.
\newblock In {\em Computer Vision and Pattern Recognition (CVPR)}, pages 1--8,
  2007.

\bibitem{SB:eccv12}
F.~Schmidt and Y.~Boykov.
\newblock Hausdorff distance constraint for multi-surface segmentation.
\newblock In {\em European Conference on Computer Vision (ECCV), LNCS 7572},
  volume~1, pages 598--611, Florence, Italy, October 2012.

\bibitem{TWRS2014}
T.~Schoenemann and V.~Kolmogorov.
\newblock Generalized sequential tree-reweighted message passing.
\newblock {\em Advanced Structured Prediction}, page~75, 2014.

\bibitem{siddiqi2008medial}
K.~Siddiqi and S.~Pizer.
\newblock {\em Medial representations: mathematics, algorithms and
  applications}, volume~37.
\newblock Springer Science \& Business Media, 2008.

\bibitem{SiddiqiPizer:2008}
K.~Siddiqi and S.~Pizer.
\newblock {\em {Medial Representations: Mathematics, Algorithms and
  Applications}}.
\newblock Springer, December 2008.

\bibitem{olga:ECCV08}
O.~Veksler.
\newblock Star shape prior for graph-cut image segmentation.
\newblock In {\em European Conference on Computer Vision (ECCV)}, 2008.

\bibitem{vu2008shape}
N.~Vu and B.~Manjunath.
\newblock Shape prior segmentation of multiple objects with graph cuts.
\newblock In {\em Computer Vision and Pattern Recognition (CVPR)}, pages 1--8,
  2008.

\end{thebibliography}
}

\end{document}